\pdfoutput=1

\documentclass[11pt]{article}

\usepackage{emnlp2021}

\usepackage{times}
\usepackage{latexsym}
\usepackage{makecell}
\usepackage{numprint}
\usepackage[T1]{fontenc}

\usepackage[utf8]{inputenc}
\usepackage{hyperref}

\usepackage{microtype}
\usepackage{multirow}
\usepackage{graphicx}
\usepackage{array}
\newcolumntype{H}{>{\setbox0=\hbox\bgroup}c<{\egroup}@{}}
\usepackage{float}
\usepackage{enumitem}

\newcounter{mycounter}
\newcommand\modelcounter{\stepcounter{mycounter}\textbf{\themycounter}}
\newcommand\setmodel[1]{\refstepcounter{mycounter}\textbf{\themycounter}\label{model:#1}}
\newcommand{\modelref}[1]{\textbf{\ref{model:#1}}}
\newcommand{\modelrefpar}[1]{(\modelref{#1})}

\newcommand{\nonen}{/~EN}
\newcommand{\fourlang}{\{EL,UK,SV,ID\}}
\newcommand{\threelang}{\{AR,RU,ZH\}}

\newcommand{\entoX}[1]{EN~$\to$~#1}
\newcommand{\nonentoX}[1]{\nonen{}~$\to$~#1}
\newcommand{\Xtoen}[1]{#1~$\to$~EN}
\newcommand{\Xtononen}[1]{#1~$\to$~\nonen{}}
\newcommand{\multialigned}[1]{#1 multi-aligned}

\title{Continual Learning in Multilingual NMT via Language-Specific Embeddings}

\author{Alexandre Bérard \\
   NAVER LABS Europe \\
   \texttt{alexandre.berard@naverlabs.com}}

\begin{document}
\maketitle
\begin{abstract}
This paper proposes a technique for adding a new source or target language to an existing multilingual NMT model without re-training it on the initial set of languages. It consists in replacing the shared vocabulary with a small language-specific vocabulary and fine-tuning the new embeddings on the new language's parallel data. Some additional language-specific components may be trained to improve performance (e.g., Transformer layers or adapter modules). Because the parameters of the original model are not modified, its performance on the initial languages does not degrade. We show on two sets of experiments (small-scale on TED Talks, and large-scale on ParaCrawl) that this approach performs as well or better as the more costly alternatives; and that it has excellent zero-shot performance: training on English-centric data is enough to translate between the new language and any of the initial languages.
\end{abstract}

\section{Introduction}

Multilingual Neural Machine Translation models are trained on multilingual data to translate from and/or into multiple languages \cite{firat-etal-2016-multi,johnson-etal-2017-googles}. Multilingual NMT is a compelling approach in production, as one only needs to train, deploy and maintain one model (instead of $2\times N$ ones, where $N$ is the number of languages). It has also been shown to improve MT quality for low-resource languages (at the cost of a slight degradation for high-resource languages) and it can allow translation between languages that have no aligned data (``zero-shot translation'').

However, such models can be costly to train, as they usually involve larger architectures and large datasets.
Moreover, because they are trained jointly on all the languages, they require to know in advance the full set of languages. Adding a new language to an existing model usually means re-training the model on the full multilingual dataset.
Naively fine-tuning the original model on the new language's data is not an option because of vocabulary mismatch (the shared vocabulary needs to be modified to include the new language's tokens) and catastrophic forgetting (the model will quickly forget how to translate in the other languages).

In this paper, we study the problem of multilingual NMT \emph{incremental training} or \emph{continual learning} and propose a novel way to efficiently add a new source or target language.

Some desirable properties of an incremental training method are:
\begin{itemize}
    \itemsep-.4em
    \item No degradation on the existing language pairs;
    \item Efficient training (e.g., no re-training on the existing language pairs);
    \item Minimal amount of added parameters: the approach should scale to many languages and the model fit on a single GPU;
    \item Minimal degradation in inference speed;
    \item Good zero-shot performance: when training with X-EN (or EN-X) data, where X is a new language, we would like the model to be able to translate from X to any known language Y (resp. from Y to X).
\end{itemize}

We propose a novel technique for incrementally adding a new source or target language, which consists in substituting the shared embedding matrix with a language-specific embedding matrix, which is fine-tuned on the new language's data only while freezing the other parameters of the model. In some cases (e.g., when the new language is on the target size), a small number of additional parameters (e.g., adapter modules) have to be trained to match the performance of the re-training baseline. We perform two sets of experiments, with a 20-language Transformer Base trained on TED Talks, and a 20-language Transformer Big (with deep encoder and shallow decoder) trained on ParaCrawl; and show that this approach is fast and parameter-efficient and that it performs as well or better as the more costly alternatives.

\section{Related work}

Some previous works study how to adapt a multilingual MT model to unseen low-resource languages, but without seeking to maintain good performance in the initial languages \cite{neubig-hu-2018-rapid,lakew2019adapting}. \citet{garcia-etal-2021-towards} introduce a ``vocabulary substitution'' approach for adding new languages to a multilingual NMT model. They create a new shared BPE vocabulary that includes the new language and initialize the embeddings of the overlapping tokens with their previous values. Then they fine-tune the entire model on the initial model's training data combined with parallel or monolingual data in the new language. Contrary to ours, their approach assumes access to the initial model's training data and results in a small performance drop in the existing languages.

\citet{lyu-etal-2020-revisiting} and \citet{escolano2020training,escolano-etal-2019-bilingual,escolano2020multilingual} propose  multi-decoder / multi-encoder architectures which they show to be compatible with incremental training. To add a new target (resp. source) language, one just has to freeze the model's encoder (resp. decoder) and train a new language-specific decoder (resp. encoder). However, this results in an enormous number of parameters.

\citet{artetxe-etal-2020-cross,pfeiffer2020unks} incrementally train language-specific embeddings for cross-lingual transfer of BERT classification models.
This approach consists of four stages: 1) train a monolingual BERT on language $L_1$; 2) train embeddings on language $L_2$ using the masked LM objective while freezing the other parameters; 3) fine-tune the $L_1$ BERT model on the desired classification task using labeled data in language $L_1$; 4) substitute the $L_1$ embeddings with the $L_2$ embeddings in the classification model and use it for $L_2$-language classification. \citet{artetxe-etal-2020-cross} also combine their approach with $L_2$-specific adapter layers and position embeddings.
While this algorithm is close to ours, it is used on encoder-only Transformers for classification tasks. Our work extends this algorithm to encoder-decoder Transformers for multilingual MT.

Also similar to our technique, \citet{thompson-etal-2018-freezing} do domain adaptation by freezing most of the NMT parameters and only fine-tuning one component (e.g., the source embeddings).
\citet{philip-etal-2020-monolingual} show that adapter modules can be used to adapt an English-centric multilingual model to unseen language pairs, but whose source and target languages are known. We wanted to go further and use adapter layers to adapt a multilingual model to unseen languages. However, we obtained the surprising result that adapting the embedding matrix is sometimes enough. In the other cases, adapter modules can be used sparingly to match baseline performance. \citet{ustun2021_denoising} introduce ``denoising adapters'' which they show can be used to incrementally adapt a multilingual MT model to new languages using monolingual data only.

\section{Techniques}

\begin{figure}
    \centering
    \includegraphics[width=.49\textwidth]{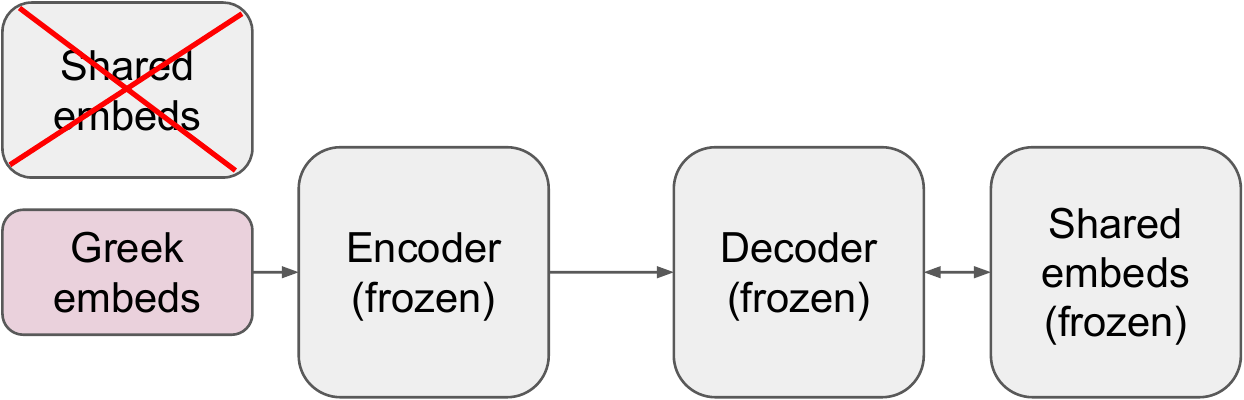}
    \caption{Adding a new source language with our incremental training technique. The source embedding matrix is replaced with the new language's embeddings and fine-tuned on the new language's data, while the other parameters are frozen.}
    \label{fig:inc_source}
\end{figure}

\begin{figure}
    \centering
    \includegraphics[width=.49\textwidth]{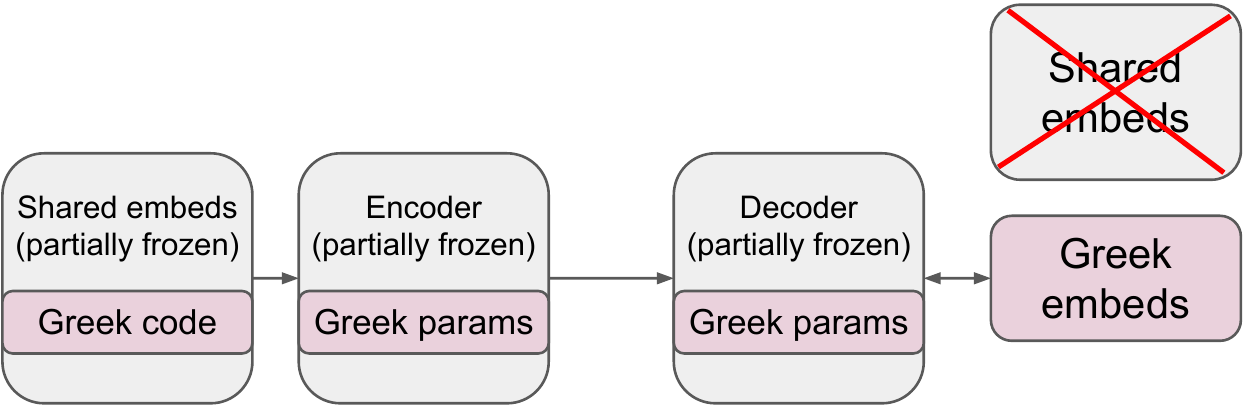}
    \caption{Adding a new target language with our incremental training technique. The tied target embedding matrix and output projection are replaced with the new language's embeddings. Some language-specific parameters can be added in the decoder or encoder, and a new language code is added in the source embedding matrix. Everything is kept frozen except for these new parameters.}
    \label{fig:inc_target}
\end{figure}

Figures~\ref{fig:inc_source} and~\ref{fig:inc_target} illustrate our technique for a new source and a new target language respectively.

The initial model is a many-to-many model with a shared vocabulary and source-side language codes (to indicate the target language).

\subsection{New source language}

To add a new source language (e.g., Greek), we build a new (smaller) vocabulary for this language only and replace the source embedding matrix with a new embedding matrix corresponding to that vocabulary. Note that some tokens may appear in both vocabularies. Similarly to \citet{pfeiffer2020unks,garcia-etal-2021-towards}, we initialize the new embeddings for those tokens with the existing embedding values.
We train this new embedding matrix on Greek-English parallel data while freezing all the other parameters.
There is no loss in performance in the existing languages as we do not modify the original parameters. At inference, to translate from the initial set of languages, we use the initial shared vocabulary and embeddings. To translate from Greek, we use the Greek embeddings and vocab.

To better adapt to the new source language, we also try combining this language-specific embedding matrix with other language-specific components in the encoder. We either fine-tune the first encoder layer while freezing the other layers, train the full encoder, or plug in adapter modules after encoder layers \cite{bapna-firat-2019-simple} and train these while freezing the Transformer parameters.

\paragraph{Data augmentation} As we will show in the experiments, source lang-specific parameters tend to give poor zero-shot results, i.e., when training them on Greek-English data, the resulting model might have trouble translating into other languages than English. For this reason, we try training such models on additional data. One solution is to use a multi-aligned Greek corpus (i.e., Greek paired with all the initial languages), but this might not always be possible. We experiment with tiny amounts of such data (e.g., 1000 line pairs per initial language); and with synthetic data: translate the English side of the Greek-English corpus into the other languages with the initial model, then use the resulting fake line pairs for training. We call this approach ``back-translation'' (BT) even though it is arguably closer to distillation than back-translation because the synthetic text will be on the target side.

\subsection{New target language}

The same incremental training techniques can be used to learn a new target language (e.g., Greek) with some modifications. The decoder has a target embedding matrix and vocabulary projection matrix, which are usually tied and shared with the source embeddings (i.e., the same parameters are used for all 3 purposes). We need to adapt both the target embeddings and output projection to the new Greek vocabulary. Like in the initial model, we tie these two parameters. Additionally, the initial model does not have a ``translate into Greek'' language code. We add this language code to the source embedding matrix and freeze all source embeddings but this one. It is initialized with the ``to English'' language code embedding of the initial model. We combine this approach with language-specific parameters (adapter modules or fine-tuned Transformer layers) in the decoder and/or encoder.

\subsection{New source and target languages}

To translate between two new languages (e.g., Greek to Ukrainian), we train language-specific parameters for each of these languages separately, as described previously. Then, at inference time, we combine these parameters. This is done by taking the new source Greek embedding matrix and target Ukrainian embedding matrix (and vocabulary projection). The ``translate into Ukrainian'' language code embedding is concatenated to the Greek embedding matrix. Similarly, the combined model includes language-specific layers and adapters from both models. When both models have adapter modules at the same layers (e.g., last encoder layer), we stack them: the target-language adapters are plugged in after the source-language adapters.

\subsection{Baselines}

We compare our incremental training techniques with two types of baselines: bilingual models trained from scratch with only the new language's parallel data; and re-training, i.e., training a new multilingual model that includes the new language. To save computation time, similarly to \citet{garcia-etal-2021-towards}, we start from the initial model and substitute its vocabulary with a new vocabulary trained with the same settings and data as before plus text in the new language. This ensures a large overlap between the old and new vocabularies. Then, we initialize the embeddings of the overlapping tokens with their previous values and fine-tune the full model on the entire dataset.

Note that these baselines do not meet our criteria for a good incremental training technique. Bilingual models are parameter-inefficient and cannot do zero-shot translation (except via pivot translation, which is twice as slow). Re-training assumes access to the initial model's training data and can be very slow. It could also result in a drop in performance in the initial languages.

\section{TED Talks Experiments}

We adapt a 20-language model trained on TED Talks to Greek (EL), either on the source side or target side. We pick Greek as the new language as it is from an unseen language family and uses an unseen alphabet. We also do experiments with Ukrainian (UK), Indonesian (ID), or Swedish (SV) as the new language,\footnote{They all use a known script (Latin or Cyrillic). Indonesian is from an unseen language family.} which are shown in Appendix.

\subsection{Data and hyper-parameters}

We use the TED Talks corpus \cite{qi-etal-2018-pre} with the same set of 20 languages as \citet{philip-etal-2020-monolingual, berard2021_efficient}.\footnote{\{en, ar, he, ru, ko, it, ja, zh\_cn, es, fr, pt\_br, nl, tr, ro, pl, bg, vi, de, fa, hu\}} This corpus is multi-parallel, i.e., it has training data for all 380 (20$\times$19) language pairs. It also includes official valid and test splits for all these language pairs. Table~\ref{tab:ted_training_data} in Appendix shows the training data size per language.

The initial model is the ``multi-parallel'' baseline from \citet{berard2021_efficient}, a Transformer Base \cite{vaswani2017attention} trained in two stages: English-centric training (38 directions) for 120 epochs; then multi-parallel fine-tuning (380 directions) for 10 epochs.\footnote{Note that an ``epoch'' when using multi-parallel data corresponds to approximately 9 English-centric epochs in terms of updates.} More hyper-parameters are given in Appendix (Table~\ref{tab:ted_hyperparameters}).

The shared vocabulary is created using BPE \cite{sennrich-etal-2016-neural} with 64k merge operations and inline casing \cite{berard-etal-2019-naver}. 
Both BPE and NMT training use temperature sampling with $T=5$ \cite{arivazhagan2019massively}. Single characters with a total frequency higher than 10 are added to the vocabulary.
The Greek vocabulary is obtained with the same BPE settings but on Greek monolingual data with 4k merge operations.
The bilingual baselines use a joint BPE model of size 8k and the same settings as in \citet{philip-etal-2020-monolingual}. Our re-training baselines are obtained by creating a new shared BPE model of size 64k including all 20 initial languages plus Greek and fine-tuning the multi-parallel model for 10 more epochs with this vocabulary. Note that there is a vocabulary mismatch with the initial model (which did not have Greek). We initialize the known embeddings with their previous values and the new ones at random and reset the learning rate scheduler and optimizer. We also do a re-training baseline that includes all 4 new languages. Note that contrary to our incremental training approach, those models are trained with the new language(s) on both sides and use multi-aligned parallel data.

Finally, we train a model that follows more closely \citet{garcia-etal-2021-towards}: we fine-tune the multi-parallel model for 10 epochs, by replacing the initial vocabulary with a vocabulary of the exact same size that includes Greek, and whose new tokens are initialized with the outdated embeddings from the old model. Like \citet{garcia-etal-2021-towards}, we upscale the new data's sampling frequency by a factor of 5.

\subsection{Evaluation settings}
\label{sec:ted_evaluation_settings}
The TED Talks models are evaluated on the provided multi-parallel validation and test sets. Since those are already word-tokenized, we run SacreBLEU with the \texttt{-{}-tok none} option.\footnote{SacreBLEU signature: \texttt{BLEU+c.mixed+\#.1+\\s.exp+tok.none+v.1.5.1}}

We report BLEU scores from/into English and average BLEU from/into the 19 other languages than English (which correspond to a zero-shot setting when the incremental training is done on Greek-English only data). We also report chrF scores obtained with SacreBLEU on the test and validation sets in Appendix.\footnote{\texttt{chrF2+numchars.6+space.false+\\version.1.5.1}}

\subsection{Results and analysis}

Table~\ref{tab:notations} in Appendix details the notations used in this paper and the tables.

\begin{table}
    \hspace{-.5cm}
    \begin{tabular}{c@{\hspace{.2cm}}|c|ccc}
    & Model & $\to$EN & $\leftarrow$EN & \nonen{} \\
    \hline
    \setmodel{bilingual} & SOTA -- bilingual & 32.4 & \textbf{24.4} & 15.0 \\
    \modelcounter{} & SOTA -- multilingual & 30.9 & 22.3 & 14.8 \\
    \hline
    \setmodel{en_centric} & English-centric & 31.8 & 24.2 & 13.5 \\
    \setmodel{multiparallel} & \modelrefpar{en_centric} + multi-parallel & 32.8 & 24.3 & 16.3 \\
    \hline
    \setmodel{retraining} & \modelrefpar{multiparallel} + EL & \textbf{33.3} & 24.3 & \textbf{16.6} \\
    \setmodel{retraining_4} & \modelrefpar{multiparallel} + \fourlang{} & 33.2 & 24.0 & 16.5 \\
    \end{tabular}
    \caption{BLEU scores (average to English, from English, and between non-English languages) of the baseline models on TED test. ``SOTA'' corresponds to the bilingual and multi-parallel baselines of \citet{philip-etal-2020-monolingual}. \modelrefpar{en_centric} and \modelrefpar{multiparallel} are from \citet{berard2021_efficient}.}
    \label{tab:ted_baselines}
\end{table}

\paragraph{Baselines.}

Table~\ref{tab:ted_baselines} compares our initial models and re-training baselines against the state of the art on the initial set of 20 languages. In this instance, fine-tuning the initial model with more languages (\modelref{retraining}, \modelref{retraining_4}) does not degrade BLEU. Appendix Table~\ref{tab:ted_baselines_chrf} shows valid and test chrF on more baselines, including our implementation of the vocabulary substitution approach of \citet{garcia-etal-2021-towards}.

\begin{table*}[t]
    \centering
        \begin{tabular}{Hc@{\hspace{.2cm}}|c|c|cc}
        & ID & Model & Params & \Xtoen{EL} & \Xtononen{EL} \\
        \cline{2-6}
        \multirow{3}{3cm}{\centering Baselines, trained with multi-aligned data} & \modelref{bilingual} & Bilingual baselines & 35.7M & 38.4 & 17.1 \\
        & \modelref{retraining} & Re-training + EL & -- & 40.4 & 19.4 \\
        & \modelref{retraining_4} & Re-training + \fourlang{} & -- & 40.2 & 19.4 \\
        \cline{2-6}
        \multirow{9}{3cm}{\centering EL-EN incremental training} & \setmodel{embed} & Only embed & 2.13M & 38.9 & 18.8 \\
        & \setmodel{embed-no-init} & \modelrefpar{embed} + random embed init & 2.13M & 38.8 & 18.7$^\star$ \\
        & \setmodel{embed+norm+bias} & \modelrefpar{embed} + enc-norm + enc-biases & 2.17M & 39.5 & 17.7 \\
        & \setmodel{enc-adapters-1} & \modelrefpar{embed} + enc-adapters-first (dim=64) & 2.19M & 39.3 & 0.6 \\
        & \setmodel{enc-adapters} & \modelrefpar{embed} + enc-adapters-all (d=64) & 2.53M & 40.3 & 0.6 \\
        & \setmodel{enc-adapters-512} & \modelrefpar{embed} + enc-adapters-all (d=512) & 5.28M & 41.0 & 0.6 \\
        & \setmodel{enc-adapters-1024-1-2-3} & \modelrefpar{embed} + enc-adapters-\{1,2,3\} (d=1024) & 5.28M & 41.0 & 0.6 \\
        & \setmodel{embed+enc-only-first} & \modelrefpar{embed} + enc-first-layer & 5.28M & 40.1 & 0.7 \\
        & \setmodel{only-enc} & \modelrefpar{embed} + all-enc-layers & 21.0M & 40.4 & 0.6 \\
        \cline{2-6}
        \multirow{6}{3cm}{\centering EL-* incremental training} & \setmodel{multi_enc-adapters-512} & \modelrefpar{enc-adapters-512} + \multialigned{EL} & 5.28M & 40.5 & \textbf{19.5} \\
        & \setmodel{multi_BT_enc-adapters-512} & \modelrefpar{enc-adapters-512} + \multialigned{EL} (BT) & 5.28M & 40.2 & 19.0 \\
        & \setmodel{enc-adapters-512+1k} & \modelrefpar{enc-adapters-512} + 1k lines per lang & 5.28M & 40.9 & 18.2 \\
        & \setmodel{enc-adapters-512+1k-BT} & \modelrefpar{enc-adapters-512} + 1k lines per lang (BT) & 5.28M & \textbf{41.2} & 17.8 \\
        & \setmodel{embed+enc-only-first+1k-BT} & \modelrefpar{embed+enc-only-first} + 1k lines per lang (BT) & 5.28M & 40.3 & 18.9 \\
        & \setmodel{enc-adapters-512+100-BT} & \modelrefpar{enc-adapters-512} + 100 lines per lang (BT) & 5.28M & 40.9 & 17.4 \\
        \cline{2-6}
        \multirow{3}{3cm}{\centering Multilingual EN-centric inc. training} & \setmodel{src4_embed} & \modelrefpar{embed} + \fourlang{} & 8.30M & 39.3 & 18.9 \\
        & \setmodel{src4_embed+enc-only-first} & \modelrefpar{embed+enc-only-first} + \fourlang{} & 11.5M & 39.9 & 2.9 \\
    \end{tabular}
    \caption{TED test BLEU scores of incremental training with Greek on the source side. ``\Xtononen{EL}'' corresponds to an average BLEU from Greek into all 19 non-English languages. ``Params'' gives the number of new parameters introduced by each approach. The initial model (\modelref{multiparallel}) has 80.2M parameters in total. ($\star$) obtained by using the ``translate into X" lang code embeddings from the initial model. The table is divided in 4 parts: baselines trained with multi-aligned data; Greek-English incremental training; incremental training with multi-aligned data (i.e., line pairs between Greek and all 20 languages); and multilingual English-centric incremental training (i.e., on 4 new source languages at once).}
    \label{tab:source_greek}
\end{table*}

\paragraph{New source language.}

Table~\ref{tab:source_greek} shows the test BLEU scores of several incrementally-trained models with Greek as a new source language. More results on Greek, Ukrainian, Indonesian and Swedish are given in Appendix (Tables~\ref{tab:source_el_chrf},~\ref{tab:source_uk_chrf},~\ref{tab:source_id_chrf}, and~\ref{tab:source_sv_chrf}).

\begin{figure*}[t]
    \centering
    \includegraphics[width=0.47\textwidth]{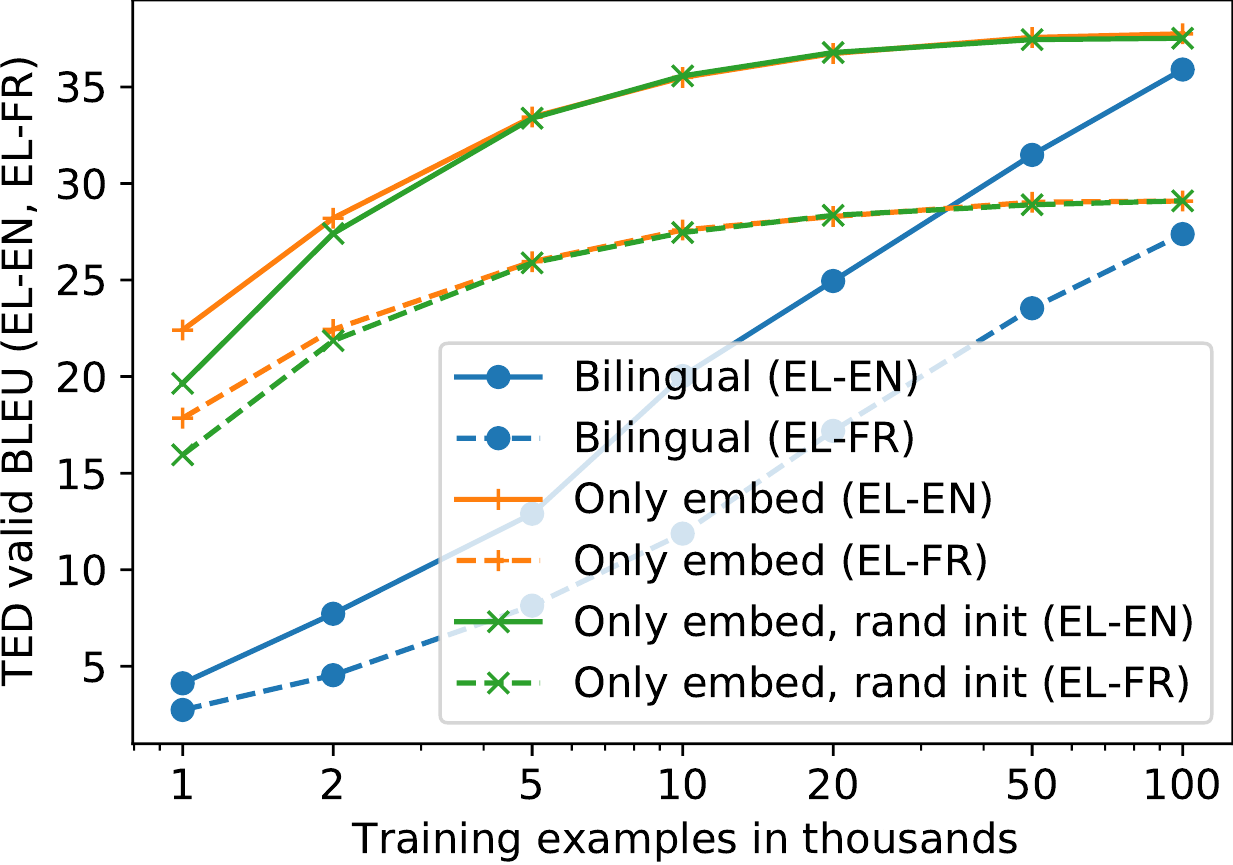}
    \hspace{.1cm}
    \includegraphics[width=0.47\textwidth]{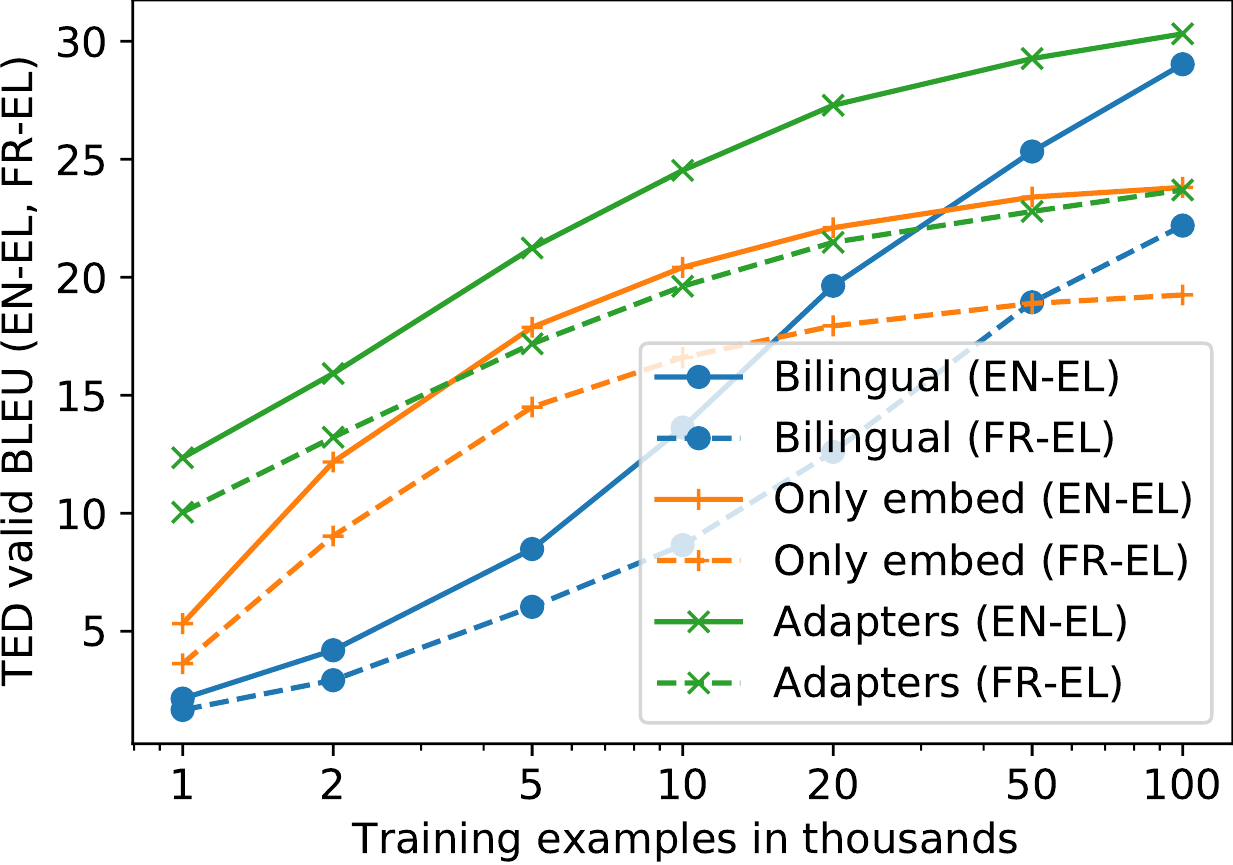}
    \caption{TED validation BLEU from Greek (left) and to Greek (right) by training corpus size, with incremental training (\modelref{embed}, \modelref{embed-no-init}, \modelref{tgt_embed+lang-code}, \modelref{tgt_embed+adapters+lang-code}) versus bilingual baselines \modelrefpar{bilingual}.}
    \label{fig:bleu_by_size}
\end{figure*}

Training the source embeddings only (\modelref{embed}) outperforms the bilingual baselines (\modelref{bilingual}) and comes close to the costly re-training baselines (\modelref{retraining}, \modelref{retraining_4}). In particular, it nearly matches the performance of the latter in the zero-shot \Xtononen{EL} directions, even though the baselines have training data for those directions. Initializing the known tokens in the new vocabulary with their old embeddings does not improve final performance (\modelref{embed} vs \modelref{embed-no-init}). But using language code embeddings from the initial model is necessary to be able to translate into non-English languages. Figure~\ref{fig:bleu_by_size} shows that such initialization improves final performance under low-resource settings. Figure~\ref{fig:ted_train_speed} in Appendix also shows that it speeds up training.

\begin{figure*}[t]
    \centering
    \includegraphics[width=0.48\textwidth]{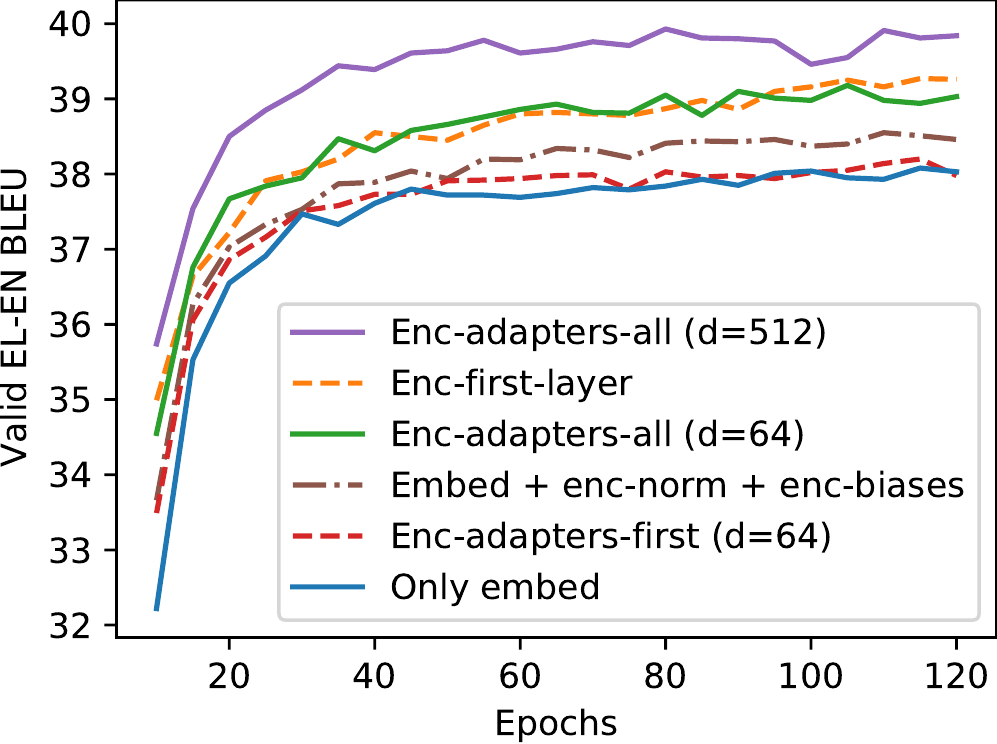}
    \includegraphics[width=0.48\textwidth]{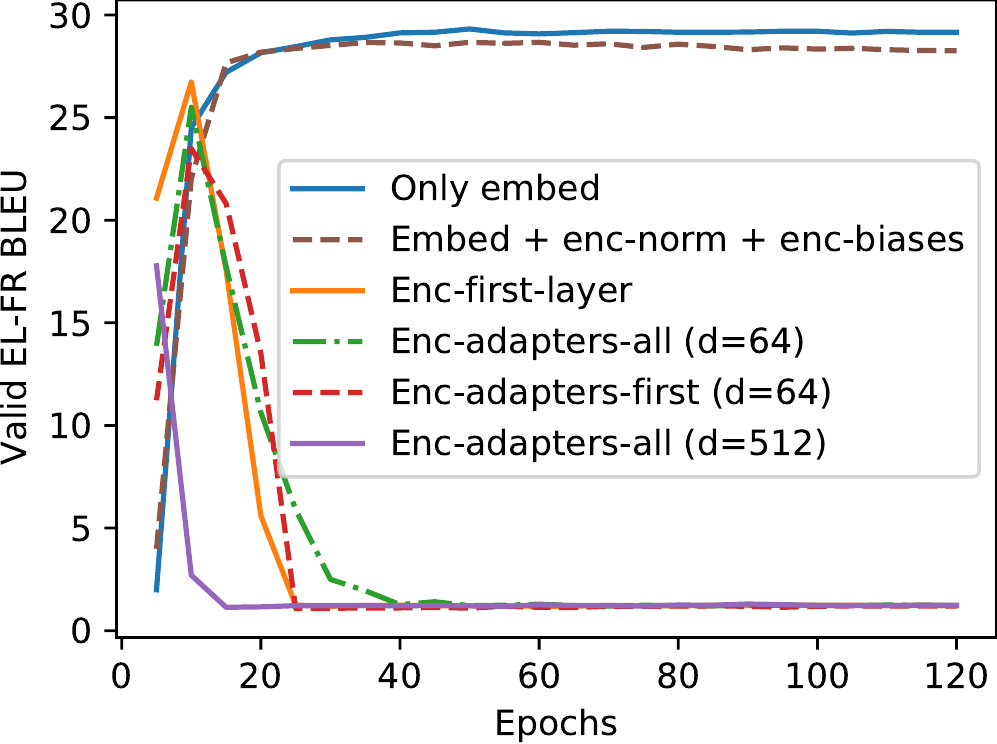}
    \caption{TED validation BLEU on EL-EN and EL-FR while incrementally training with EL-EN data only (\modelref{embed}, \modelref{embed+norm+bias}, \modelref{enc-adapters-1}, \modelref{enc-adapters},  \modelref{enc-adapters-512}, \modelref{embed+enc-only-first}).}
    \label{fig:catastrophic_forgetting}
\end{figure*}

Training additional components in the encoder, like adapter modules (\modelref{enc-adapters}, \modelref{enc-adapters-512}, \modelref{enc-adapters-1024-1-2-3}) or the first encoder layer (\modelref{embed+enc-only-first}) helps improve \Xtoen{EL} performance and outperform all baselines, though it is less useful when the new language is from a known family (see Ukrainian and Swedish scores in Appendix Tables~\ref{tab:source_uk_chrf} and~\ref{tab:source_sv_chrf}). However, this results in abysmal zero-shot performance (\Xtononen{EL}). As they only encounter the ``to English'' language code during training, those models quickly forget how to interpret the other lang codes. This catastrophic forgetting is illustrated by Figure~\ref{fig:catastrophic_forgetting}, where we see a plunge in EL~$\to$~FR performance after just a few epochs of training. Only tuning the encoder layer norm parameters and biases (\modelref{embed+norm+bias}) gives slightly higher \Xtoen{EL} performance without suffering from catastrophic forgetting in the other languages.
Note that language code forgetting is less pronounced when the initial model is English-centric (see Table~\ref{tab:source_greek_en} in Appendix). In this setting, adapter modules do not hurt zero-shot translation.

\begin{table*}[t]
    \hspace{-.3cm}
    \begin{tabular}{c@{\hspace{.2cm}}|c|c|cc}
        ID & Model & Params & \entoX{EL} & \nonentoX{EL} \\
        \hline
        \modelref{bilingual} & Bilingual baselines & 35.7M & 32.2 & 18.3 \\
        \modelref{retraining} & Re-training + EL & -- & 32.5 & \textbf{21.1} \\
        \modelref{retraining_4} & Re-training + \fourlang{} & -- & 32.1 & \textbf{21.1} \\
        \hline
        \setmodel{tgt_embed+lang-code} & Only embed & 2.13M & 25.7 & 16.7 \\
        \setmodel{tgt_embed+non-tied+lang-code} & \modelrefpar{tgt_embed+lang-code} + non-tied & 4.25M & 27.1 & 17.8 \\
        \setmodel{tgt_embed+dec-adapters+lang-code} & \modelrefpar{tgt_embed+lang-code} + dec-adapters-all (dim=64) & 2.53M & 29.8 & 19.3 \\
        \setmodel{tgt_embed+adapters+lang-code} & \modelrefpar{tgt_embed+lang-code} + adapters-all (d=64) & 2.93M & 32.7 & 19.6 \\
        \setmodel{tgt_embed+enc-adapters-6-1024+lang-code} & \modelrefpar{tgt_embed+lang-code} + enc-adapters-last (d=1024) & 3.18M & 32.0 & 20.0 \\
        \setmodel{tgt_embed+enc-adapters-6-1024+dec-adapters+lang-code} & \modelrefpar{tgt_embed+dec-adapters+lang-code} + enc-adapters-last (d=1024) & 3.58M & 33.5 & 20.6 \\
        \setmodel{tgt_embed+dec-only-last+lang-code} & \modelrefpar{tgt_embed+lang-code} + dec-last-layer & 6.33M & 32.6 & 20.5 \\
        \setmodel{tgt_embed+dec-only-last+enc-adapters-6-1024+lang-code} & \modelrefpar{tgt_embed+dec-only-last+lang-code} + enc-adapters-last (d=1024) & 7.38M & 34.0 & 20.8 \\
        \setmodel{tgt_embed+adapters-430+lang-code} & \modelrefpar{tgt_embed+lang-code} + adapters-all (d=430) & 7.43M & 34.0 & 18.2 \\
        \setmodel{tgt_embed+enc-adapters-6-1024+dec-adapters-690+lang-code} & \modelrefpar{tgt_embed+lang-code} + dec-adapters-all (d=690) + enc-adapters-last (d=1024) & 7.43M & 33.8 & 20.8 \\
        \setmodel{tgt_embed+dec-only-last+adapters-90+lang-code} & \modelrefpar{tgt_embed+dec-only-last+lang-code} + adapters-all (d=90) & 7.36M & \textbf{34.2} & 19.8 \\
        \setmodel{tgt_embed+dec-only-last+enc-adapters-170+lang-code} & \modelrefpar{tgt_embed+dec-only-last+lang-code} + enc-adapters-all (d=170) & 7.38M & 34.1 & 19.0 \\
        \hline
        \setmodel{tgt_multi_embed+dec-only-last+enc-adapters-6-1024+lang-code} & \modelrefpar{tgt_embed+dec-only-last+enc-adapters-6-1024+lang-code} + \multialigned{EL} & 7.35M & 32.9 & \textbf{21.1} \\
        \hline
        \setmodel{tgt_inc4_embed+dec-only-last+enc-adapters-6-1024+lang-code} & \modelrefpar{tgt_embed+dec-only-last+enc-adapters-6-1024+lang-code} + \fourlang{} & 13.5M & 33.0 & 20.4 \\
    \end{tabular}
    \caption{TED test BLEU scores of incremental training with Greek on the target side. ``\nonentoX{EL}'' corresponds to an average BLEU from the 19 non-English languages to Greek. ``Params'' gives the number of new parameters introduced by each approach.}
    \label{tab:target_greek}
\end{table*}

\begin{table*}[t]
    \centering
    \begin{tabular}{Hcc|cc|c}
        ID & & Source model & & Target model & BLEU \\
        \hline
        & \modelref{bilingual} & \multicolumn{3}{c|}{Bilingual} & 14.9 \\
        & \modelref{bilingual} & \multicolumn{3}{c|}{Bilingual (pivot through English)} & 18.5 \\
        & \modelref{retraining_4} & \multicolumn{3}{c|}{Re-training + \fourlang{}} & \textbf{22.0} \\
        \hline
        & \multirow{2}{*}{\modelref{embed}} & \multirow{2}{*}{Only embed} & \modelref{tgt_embed+dec-only-last+lang-code} & Dec-last-layer & 21.1 \\
        & & & \modelref{tgt_embed+dec-only-last+enc-adapters-6-1024+lang-code} & Dec-last-layer + enc-adapters-last (d=1024) & 21.0 \\
        \hline
        & \multirow{3}{*}{\modelref{embed+enc-only-first}} & \multirow{3}{*}{Enc-first-layer} & \modelref{tgt_embed+dec-only-last+lang-code} & Dec-last-layer & 20.7 \\
        & & & \modelref{tgt_embed+dec-only-last+enc-adapters-6-1024+lang-code} & Dec-last-layer + enc-adapters-last (d=1024) & 21.3  \\
        & & & \modelref{tgt_embed+dec-only-last+enc-adapters-6-1024+lang-code} & Pivot through English$^\star$ & 21.6 \\
        \hline
        & \multirow{2}{*}{\modelref{embed+enc-only-first+1k-BT}} & \multirow{2}{*}{Enc-first-layer + 1k (BT)} & \modelref{tgt_embed+dec-only-last+lang-code} & Dec-last-layer & 21.2 \\
        & & & \modelref{tgt_embed+dec-only-last+enc-adapters-6-1024+lang-code} & Dec-last-layer + enc-adapters-last (d=1024) & 21.3 \\
        \hline
        & \multirow{2}{*}{\modelref{enc-adapters-512+1k-BT}} &  \multirow{2}{*}{Enc-adapters-all (d=512) + 1k (BT)} & \modelref{tgt_embed+dec-only-last+lang-code} & Dec-last-layer & 20.4 \\
        & & & \modelref{tgt_embed+dec-only-last+enc-adapters-6-1024+lang-code} & Dec-last-layer + enc-adapters-last (d=1024) & 20.7 \\
    \end{tabular}
    \caption{TED test BLEU scores on \fourlang{}$\to$\fourlang{} (average over 12 directions) by combining source-language and target-language incrementally-trained parameters. ($\star$) instead of combining model parameters, translate to English with \modelrefpar{embed+enc-only-first}, then to the target language with \modelrefpar{tgt_embed+dec-only-last+enc-adapters-6-1024+lang-code}.}
    \label{tab:source_target_inc}
\end{table*}

The third quarter of Table~\ref{tab:source_greek} shows how multi-aligned Greek data can be used to achieve excellent performance in both \Xtoen{EL} and \Xtononen{EL} directions.
The best tradeoff between \Xtoen{EL} and \Xtononen{EL} performance is achieved by incrementally training with the entire Greek dataset of 2.41M line pairs. However, such data might not always be accessible for the new language. Close performance can be reached by training with the same amounts of synthetic data instead (\modelref{multi_BT_enc-adapters-512}). And more interestingly, only a tiny amount of real (\modelref{enc-adapters-512+1k}) or back-translated data (\modelref{enc-adapters-512+1k-BT}, \modelref{embed+enc-only-first+1k-BT}, \modelref{enc-adapters-512+100-BT}) in the other 19 languages is needed to obtain good zero-shot results without any loss in \Xtoen{EL} performance.

\paragraph{New target language.}
Table~\ref{tab:target_greek} shows test BLEU scores when incrementally adding Greek on the target side. Additional results on Greek, Ukrainian, Indonesian and Swedish are provided in Appendix (Tables~\ref{tab:target_el_chrf},~\ref{tab:target_uk_chrf},~\ref{tab:target_id_chrf}, and~\ref{tab:target_sv_chrf}).

With new target languages, only adapting the embedding matrix (tied with vocabulary projection) is not enough and strongly underperforms the baselines (\modelref{tgt_embed+lang-code} vs \modelref{bilingual}, \modelref{retraining} and \modelref{retraining_4}). Training decoder-side adapter modules (\modelref{tgt_embed+dec-adapters+lang-code}) gets us closer to baseline performance; and tuning the last decoder layer (\modelref{tgt_embed+dec-only-last+lang-code}) bridges the gap with the baselines. However, the most effective strategy is to train some components in both the encoder and decoder (\modelref{tgt_embed+adapters+lang-code}, \modelref{tgt_embed+enc-adapters-6-1024+dec-adapters+lang-code}, \modelref{tgt_embed+dec-only-last+enc-adapters-6-1024+lang-code}, \modelref{tgt_embed+adapters-430+lang-code}, \modelref{tgt_embed+enc-adapters-6-1024+dec-adapters-690+lang-code}, \modelref{tgt_embed+dec-only-last+adapters-90+lang-code}, \modelref{tgt_embed+dec-only-last+enc-adapters-170+lang-code}).
We observed that it was important for the model to have a way to modify the output of the encoder before it is read by frozen decoder components. Interestingly, only having a large adapter module after the last encoder layer (\modelref{tgt_embed+enc-adapters-6-1024+lang-code}) is enough to match baseline performance. Adding small adapters after each decoder layer (\modelref{tgt_embed+enc-adapters-6-1024+dec-adapters+lang-code}) further improves BLEU and brings the best parameter count / performance tradeoff.

At the same parameter budget, training adapter modules after every encoder layer (\modelref{tgt_embed+adapters-430+lang-code}, \modelref{tgt_embed+dec-only-last+adapters-90+lang-code}, \modelref{tgt_embed+dec-only-last+enc-adapters-170+lang-code}) gives worse \nonentoX{EL} performance than an adapter at the last encoder layer combined with decoder-side parameters (\modelref{tgt_embed+dec-only-last+enc-adapters-6-1024+lang-code}, \modelref{tgt_embed+enc-adapters-6-1024+dec-adapters-690+lang-code}), which is likely caused by the encoder overfitting to English.

In this setting, there is no clear advantage to incremental training with multi-aligned Greek data (\modelref{tgt_multi_embed+dec-only-last+enc-adapters-6-1024+lang-code}), as this hurts \entoX{EL} performance, without any notable improvement for \nonentoX{EL}. Finally, multilingual incremental training (with 4 new target languages at once) is entirely possible (\modelref{tgt_inc4_embed+dec-only-last+enc-adapters-6-1024+lang-code}) and gives competitive results to the baselines.

Table~\ref{tab:lang_code_ablation_study} in Appendix analyzes the usefulness of learning a new language code, by comparing with three other strategies: incremental training without any language code; with the ``to English'' language code; or with the language code of a similar language. Interestingly, the more new parameters are learned (esp. encoder-side), the less useful it is to learn a new language code. Moreover, adapting to Swedish by using a fixed English language code gives reasonable performance as the two languages are from the same family. And the proxy ``to Russian'' language code gives the same results as learning a new language code when adapting to Ukrainian because both languages are very similar.

\paragraph{New source and target languages}

Table~\ref{tab:source_target_inc} combines incrementally-trained parameters at inference time to translate between two new languages. Interestingly, combining target-language parameters with source-language parameters that had very poor zero-shot performance (\modelref{embed+enc-only-first}) gives excellent results. We hypothesize that the language code forgetting issue is less pronounced here because solely activating some language-specific parameters will make the model translate into that language.

Despite showing the best \Xtoen{EL} performance, source-language encoder adapters (\modelref{enc-adapters-512+1k-BT}) tend to perform more poorly when combined with target-language parameters. 
While better performance is obtained by pivot translation through English with two incrementally trained models (\modelref{embed+enc-only-first} and \modelref{tgt_embed+dec-only-last+enc-adapters-6-1024+lang-code}), combining the parameters of these two models gives close results at a faster inference speed.

\paragraph{Discussion}

Figure~\ref{fig:bleu_by_size} shows final BLEU scores of our techniques when training with smaller amounts of data. We observe that incremental training is more data-efficient than bilingual models and can achieve decent performance even with tiny amounts of training data, making it a good fit for adaptation to low-resource languages.

Figure~\ref{fig:ted_train_speed} in Appendix illustrates the training speed of our approach compared to our implementation of \citet{garcia-etal-2021-towards}. In addition to maintaining the exact same performance on the previous languages and needing only English-centric data, our approach reaches higher performance in much fewer updates than the alternatives. Note that re-training might be an efficient solution if one wants to add several languages at once and on both sides.

Finally, Tables~\ref{tab:source_greek_en} and~\ref{tab:target_greek_en} in Appendix show that our incremental training approach can be applied to English-centric initial models with similar results.

\section{ParaCrawl Experiments}

\begin{table}
    \hspace{-.3cm}
    \begin{tabular}{c@{\hspace{.2cm}}|c|ccc}
    ID & Model & $\to$EN & $\leftarrow$EN & \nonen{} \\
    \hline
    \setmodel{paracrawl_sota} & M2M-124 & 32.4 & 31.9 & 25.7 \\
    \hline
    \setmodel{paracrawl_en_centric_big_6_6} & Big 6-6 EN-centric & 38.8 & 36.4 & 18.5 \\
    \setmodel{paracrawl_en_centric} & Big 12-2 EN-centric & \textbf{39.6} & \textbf{37.1} & 21.1 \\
    \setmodel{paracrawl_en_centric_pivot} & \modelrefpar{paracrawl_en_centric} + pivot (EN) & -- & -- & \textbf{27.6} \\
    \setmodel{paracrawl_multiparallel} & \modelrefpar{paracrawl_en_centric} + multi-para. & 39.0	& 36.2 & \textbf{27.6} \\
    \hline
    \setmodel{paracrawl_en_centric_retraining} & \modelrefpar{paracrawl_en_centric} + \threelang{} & \textbf{39.6} & 36.6 & 21.0 \\
    \setmodel{paracrawl_retraining} & \modelrefpar{paracrawl_multiparallel} + \threelang{} & 39.0 & 35.8 & 27.5 \\
    \end{tabular}
    \caption{FLORES devtest spBLEU scores of the ParaCrawl/UNPC baselines. Average to English (19 directions), from English (19 directions) and between non-English languages (342 directions). (\modelref{paracrawl_en_centric_big_6_6}, \modelref{paracrawl_en_centric}, \modelref{paracrawl_en_centric_pivot}, \modelref{paracrawl_multiparallel}) are the same models as in \cite{berard2021_efficient}.}
    \label{tab:paracrawl_baselines}
\end{table}

In this section, we test our approach in a more realistic, large-scale setting: a 20-language Transformer Big initial model trained on ParaCrawl \cite{banon-etal-2020-paracrawl}.
The incremental training experiments are done in three languages: Arabic (AR), Russian (RU), and Chinese (ZH). Arabic and Chinese are both from unseen language families and use unseen scripts. Russian is close to a known language (Bulgarian) and uses the same script. For training on those languages, we use UNPC \cite{ziemski-etal-2016-united}.
\newcolumntype{P}[1]{>{\centering\arraybackslash}p{#1}}
\begin{table*}[t]
    \hspace{-.2cm}
    \begin{tabular}{c@{\hspace{.2cm}}|P{6.75cm}|c|ccc|ccc}
        \multirow{2}{*}{ID} & \multirow{2}{*}{Model} & \multirow{2}{*}{Params} & AR & RU & ZH & AR & RU & ZH \\
        & & & \multicolumn{3}{c|}{$\to$ EN} & \multicolumn{3}{c}{$\to$ \nonen{}} \\
        \hline
        \modelref{paracrawl_sota} & M2M-124 \cite{goyal2021flores101} & -- & 25.5 & 27.5 & 20.9 & 19.6 & 24.0 & 18.7 \\
        \setmodel{paracrawl_bilingual} & Bilingual baselines (pivot through English) & 193M & \textbf{32.1} & 29.9 & \textbf{23.7} & \textbf{23.2} & 23.9 & 19.6 \\
        \modelref{paracrawl_en_centric_retraining} & English-centric \modelrefpar{paracrawl_en_centric} + \threelang{} & -- & 30.0 & \textbf{31.7} & 22.6 & 14.6 & 16.9 & 11.8 \\
        \modelref{paracrawl_retraining} & Multi-parallel \modelrefpar{paracrawl_multiparallel} + \threelang{} & -- & 31.0 & 31.6 & 23.1 & 19.2 & 23.5 & 15.9 \\
        \modelref{paracrawl_retraining_pivot} & \modelrefpar{paracrawl_retraining} + pivot through English & -- & -- & -- & -- & 22.2 & 24.5 & \textbf{19.1} \\
        \hline
        \setmodel{para_src_embed} & Only embed & 8.6M & 24.2 & 30.8 & 20.5 & 16.5 & 23.8 & 15.6 \\
        \setmodel{para_src_embed+adapters_512} & \modelrefpar{para_src_embed} + enc-adapters-all (dim=512) & 21.3M & 31.7 & 31.3 & 23.5 & 1.0 & 1.5 & 1.5 \\
        \setmodel{para_src_embed+layer-1} & \modelrefpar{para_src_embed} + enc-first-layer & 21.2M & 30.3 & 31.2 & 23.2 & 1.0 & 2.6 & 1.5 \\
        \setmodel{para_src_embed+layer-1_multipara} & \modelrefpar{para_src_embed+layer-1} + 20k lines per lang (BT) & 21.2M & 30.1 & 31.4 & 22.8 & 20.8 & \textbf{24.6} & 17.8 \\
    \end{tabular}
    \\[.2cm]
    
    \hspace{-.2cm}
    \begin{tabular}{c@{\hspace{.2cm}}|c|c|ccc|ccc}
        \multirow{2}{*}{ID} & \multirow{2}{*}{Model} & \multirow{2}{*}{Params} & \multicolumn{3}{c|}{EN $\to$} & \multicolumn{3}{c}{\nonen{} $\to$} \\
        & & & AR & RU & ZH & AR & RU & ZH \\
        \hline
        \modelref{paracrawl_sota} & M2M-124 \cite{goyal2021flores101} & -- & 17.9 & 27.1 & 19.3 & 13.8 & \textbf{23.0} & 16.6 \\
        \modelref{paracrawl_bilingual} & Bilingual baselines (pivot through English) & 193M & \textbf{29.1} & 27.5 & \textbf{22.9} & \textbf{21.2} & 22.5 & \textbf{18.2} \\
        \modelref{paracrawl_en_centric_retraining} & English-centric \modelrefpar{paracrawl_en_centric} + \threelang{} & -- & 26.5 & 26.8 & 18.4 & 15.7 & 19.3 & 11.3 \\
        \modelref{paracrawl_retraining} & Multi-parallel \modelrefpar{paracrawl_multiparallel} + \threelang{} & -- & 28.3 & 27.3 & 20.6 & 16.6 & 21.6 & 13.1 \\
        \setmodel{paracrawl_retraining_pivot} & \modelrefpar{paracrawl_retraining} + pivot through English & -- & -- & -- & -- & 20.2 & 21.8 & 16.3 \\
        \hline
        \setmodel{para_tgt_embed} & Only embed & 8.6M & 11.6 & 19.7 & 14.0 & 9.3 & 15.8 & 11.5 \\
        \setmodel{para_tgt_adapters} & \modelrefpar{para_tgt_embed} + enc-adapt-last + dec-adapt-all (1024) & 14.9M & 27.0 & 26.2 & 20.9 & 18.8 & 20.8 & 16.8 \\
        \setmodel{para_tgt_last_layer} & \modelrefpar{para_tgt_embed} + dec-last-layer & 25.4M & 26.5 & 26.9 & 21.5 & 19.1 & 20.9 & 17.4 \\
        \setmodel{para_tgt_last_layer+enc_adapter} & \modelrefpar{para_tgt_last_layer} + enc-adapters-last (dim=1024) & 27.5M & 28.2 & 28.0 & 22.5 & 20.0 & 21.8 & 18.0 \\
        \setmodel{para_tgt_last_layer+enc_adapter_no_lang_id} & \modelrefpar{para_tgt_last_layer+enc_adapter} without lang ID filtering & 27.5M & 18.0 & 19.0 & 13.6 & 5.2 & 10.3 & 5.0 \\
        \setmodel{para_tgt_full_dec} & \modelrefpar{para_tgt_embed} + all-dec-layers & 42.2M & 28.6 & \textbf{28.1} & 22.4 & 20.1 & 21.9 & 18.1 \\
    \end{tabular}
    \caption{FLORES devtest spBLEU scores of the ParaCrawl/UNPC incrementally-trained models. The top half of each table corresponds to the baselines (SOTA, bilingual or re-training). ``Params'' gives the number of new parameters introduced by each approach. The incremental training is always done on one language only (i.e., one row can correspond to 3 different models). Note that the parameter counts given in this table are for Arabic (8.63M embedding parameters). Russian and Chinese embeddings have respectively 8.51M and 13.60M parameters.}
    \label{tab:paracrawl_inc}
\end{table*}

\subsection{Data and hyper-parameters}

We download ParaCrawl v7.1 in the 19 highest-resource languages paired with English.\footnote{\{fr, de, es, it, pt, nl, nb, cs, pl, sv, da, el, fi, hr, hu, bg, ro, sk, lt\}} Then, like \citet{freitag-firat-2020-complete}, we build a multi-parallel corpus by aligning all pairs of languages through their English side (effectively removing any English duplicate). See Appendix Table~\ref{tab:paracrawl_training_data} for training data statistics.
We create a shared BPE vocab with 64k merge operations and inline casing, by sampling from this data with $T=5$ and include all characters whose frequency is higher than 100.

Our initial model is the Transformer Big 12-2 (i.e., with a deep encoder and shallow decoder) multi-parallel model of \citet{berard2021_efficient}. Like in the previous section, it was trained in two stages:
English-centric training (with $T=5$) for 1M steps; then multi-parallel fine-tuning (with $T=2$) for 200k more steps. 
More hyper-parameters are given in Appendix (Table~\ref{tab:paracrawl_hyperparameters}).

Incremental training is done for 120k steps with English-centric data from UNPC v1.0 (see Table~\ref{tab:unpc_training_data} in Appendix for statistics), which we clean by removing any line pairs where either side is detected as being in the wrong language by \texttt{langid.py} \cite{lui-baldwin-2012-langid}. We use monolingual BPE models of size 8k. The Chinese data is tokenized with Jieba\footnote{\url{https://github.com/fxsjy/jieba}} before learning the BPE model.

The English-centric bilingual baselines are Transformer Big 6-6 models trained on UNPC for 120k steps with joint BPE vocabularies of size 16k. We do two ``re-training'' baselines, by fine-tuning either the English-centric or multi-parallel model on their initial ParaCrawl data plus UNPC data in all three new languages. We sample UNPC line pairs in each of the new directions with probability $0.05$. The remaining $0.7$ probability mass is distributed among the initial ParaCrawl directions with $T=5$. The new BPE vocabulary is trained with the monolingual ParaCrawl/UNPC data in all 23 languages (with $T=5$). The new embeddings for the known tokens are initialized with their old values and the other embeddings at random.\footnote{78\% of the new tokens were in the initial vocabulary, and 84\% of the old tokens are in the new vocabulary.} Note that contrary to the TED Talks experiments, we do not have multi-aligned data for the new languages.

\subsection{Evaluation settings}
For validation, we use our own split from TED2020 \cite{reimers-gurevych-2020-making}: 3000 random line pairs for each translation direction. We report chrF scores\footnote{\url{chrF2+n.6+s.false+v.1.5.1}} computed on these valid sets in Appendix.

As test sets, we use FLORES devtest \cite{goyal2021flores101}. We report scores computed with their new ``spBLEU'' metric,\footnote{\texttt{BLEU+c.mixed+\#.1+s.exp+tok.spm+v.1.5.1} (\url{https://github.com/ngoyal2707/sacrebleu})} which runs BLEU on top of a standardized multilingual BPE tokenization.

\subsection{Results and analysis}

\paragraph{Baselines.}
Table~\ref{tab:paracrawl_baselines} compares our initial model \modelrefpar{paracrawl_multiparallel} with other baselines. Our multi-parallel model beats the M2M-124 model of \citet{goyal2021flores101} in all three settings. This is not so surprising, as their model only has 615M parameters for 124 languages, compared to 255M parameters for our 20-language model. 
Last, we can observe that our ``re-training'' baselines (\modelref{paracrawl_en_centric_retraining} and \modelref{paracrawl_retraining}) perform almost as well as the initial 20-language models (\modelref{paracrawl_en_centric}, \modelref{paracrawl_multiparallel}).

\begin{table*}
    \centering
    \begin{tabular}{Hcc|cc|c}
        ID & & Source model & & Target model & spBLEU \\
        \hline
        & \modelref{paracrawl_sota} & \multicolumn{3}{c|}{M2M-124 \cite{goyal2021flores101}} & 15.0 \\
        & \modelref{paracrawl_bilingual} & \multicolumn{3}{c|}{Bilingual baselines (pivot through English)} & \textbf{18.3} \\
        & \modelref{paracrawl_retraining} & \multicolumn{3}{c|}{Multi-parallel \modelrefpar{paracrawl_multiparallel} + \threelang{}} & 11.3 \\
        & \modelref{paracrawl_retraining_pivot} & \multicolumn{3}{c|}{\modelrefpar{paracrawl_retraining} + pivot through English} & 16.8 \\
        \hline
        & \multirow{2}{*}{\modelref{para_src_embed}} & \multirow{2}{*}{Only embed} & \modelref{para_tgt_last_layer} & Dec-last-layer & 13.5 \\
        & & & \modelref{para_tgt_last_layer+enc_adapter} & Dec-last-layer + enc-adapters-last (d=1024) & 13.9 \\
        \hline
        & \multirow{3}{*}{\modelref{para_src_embed+adapters_512}} &  \multirow{3}{*}{Enc-adapters-all (dim=512)} & \modelref{para_tgt_last_layer} & Dec-last-layer & 13.0 \\
        & & & \modelref{para_tgt_last_layer+enc_adapter} & Dec-last-layer + enc-adapters-last (d=1024) & 13.8 \\
        & & & \modelref{para_tgt_last_layer+enc_adapter} & Pivot through English & 17.7 \\
        \hline
        & \multirow{2}{*}{\modelref{para_src_embed+layer-1}} & \multirow{2}{*}{Enc-first-layer} & \modelref{para_tgt_last_layer} & Dec-last-layer & 10.2 \\
        & & & \modelref{para_tgt_last_layer+enc_adapter} & Dec-last-layer + enc-adapters-last (d=1024) & 10.6 \\
        \hline
        & \multirow{2}{*}{\modelref{para_src_embed+layer-1_multipara}} & \multirow{2}{*}{Enc-first-layer + 20k (BT)} & \modelref{para_tgt_last_layer} & Dec-last-layer & 12.0 \\
        & & & \modelref{para_tgt_last_layer+enc_adapter} & Dec-last-layer + enc-adapters-last (d=1024) & 12.7 \\
    \end{tabular}
    \caption{FLORES devtest spBLEU scores of the ParaCrawl/UNPC models on \threelang{}$\to$\threelang{} (average over 6 directions) by combining source-language and target-language incrementally-trained parameters.}
    \label{tab:paracrawl_source_target_inc}
\end{table*}

\paragraph{New source or target language.}
Training only source embeddings (\modelref{para_src_embed}) is a good strategy for Russian, but underperforms the baselines in the more linguistically distant Arabic and Chinese. Learning more parameters (+8\% per source language) can match baseline performance in all 3 languages (\modelref{para_src_embed+adapters_512} and \modelref{para_src_embed+layer-1}), but gives poor zero-shot performance. Adding small amounts of ``back-translated'' data (\modelref{para_src_embed+layer-1_multipara}) achieves close non-English performance to the pivot translation baselines without hurting English-centric scores. For new target languages, the best strategy is to train the last decoder layer with an adapter module at the last encoder layer (\modelref{para_tgt_last_layer+enc_adapter}), which matches the re-training baselines in all 3 languages and gets close performance to the parameter-inefficient bilingual baselines.
Interestingly, target-side incremental training is very sensitive to training data noise. In a first iteration of our experiments, we trained with unfiltered UNPC data and observed catastrophic performance (\modelref{para_tgt_last_layer+enc_adapter_no_lang_id}). Simple language ID filtering solved this issue.

\paragraph{New source and target languages.}

Table~\ref{tab:paracrawl_source_target_inc} combines source-language with target-language incrementally-trained parameters to translate between two new languages. The results are not as good as in our TED Talks experiments. The best combination in this setting (\modelref{para_src_embed} with \modelref{para_tgt_last_layer+enc_adapter}) performs considerably worse than pivot translation through English with the baselines. However, it outperforms the ``re-training'' baseline \modelrefpar{paracrawl_retraining}, which has only seen English-centric data for the new languages. And pivot translation with two incrementally-trained models (\modelref{para_src_embed+adapters_512} with \modelref{para_tgt_last_layer+enc_adapter}) gives excellent results, close to the bilingual baselines.

\section{Conclusion}

We propose a new technique for incrementally training multilingual NMT models on a new source or target language. It consists in creating a new monolingual BPE vocabulary for that language, substituting the shared embedding matrix with language-specific embeddings, and training those while freezing the other model parameters. At inference, translating in any of the initial languages is done by using the initial shared embeddings, and translating in the new language is done by using the newly trained embeddings. This approach does not change performance on the initial languages as the initial parameters are kept aside and not modified. For new source languages, it can achieve close performance to the more costly and less flexible bilingual and re-training baselines. For new target languages, this technique can be combined with language-specific parameters (fine-tuned Transformer layers or adapter modules) to match baseline performance at a small parameter cost.
We validate this technique on two sets of experiments: small-scale on TED Talks and large-scale on ParaCrawl; and show that it is compatible with two architectures: Transformer Base 6-6 and Big 12-2. We also show that incremental training on data aligned with English is enough to learn to translate between the new language and any of the initial languages. Translation between a new source and a new target language is also possible by combining their respective parameters at inference.

\clearpage

\bibliography{main}
\bibliographystyle{acl_natbib}

\clearpage
\appendix
\onecolumn

\begin{table*}[t]
    \section{Appendix}
    \vspace{.5cm}
	\centering
	\begin{tabular}{c|c|c|c|c|c}
    Language & Code & Family & Script & X-EN lines & X-* lines \\
    \hline
    English & en & Germanic & Latin & 3.56M & 3.56M \\
    Arabic & ar & Semitic & Arabic & 214.1k & 3.43M \\
    Hebrew & he & Semitic & Hebrew & 211.8k & 3.40M \\
    Russian & ru & Slavic & Cyrillic & 208.5k & 3.38M \\
    Korean & ko & Koreanic & Hangul & 205.6k & 3.35M \\
    Italian & it & Romance & Latin & 204.5k & 3.35M \\
    Japanese & ja & Japonic & Chinese + Kana & 204.1k & 3.31M \\
    Mandarin Chinese & zh\_cn & Sinitic & Chinese & 199.9k & 3.30M \\
    Spanish & es & Romance & Latin & 196.0k & 3.23M \\
    French & fr & Romance & Latin & 192.3k & 3.19M \\
    Brazilian Portuguese & pt\_br & Romance & Latin & 184.8k & 3.11M \\
    Dutch & nl & Germanic & Latin & 183.8k & 3.05M \\
    Turkish & tr & Turkic & Latin & 182.5k & 3.02M \\
    Romanian & ro & Romance & Latin & 180.5k & 3.06M \\
    Polish & pl & Slavic & Latin & 176.2k & 3.00M \\
    Bulgarian & bg & Slavic & Cyrillic & 174.4k & 2.95M \\
    Vietnamese & vi & Vietic & Latin & 172.0k & 2.81M \\
    German & de & Germanic & Latin & 167.9k & 2.90M \\
    Persian & fa & Iranian & Arabic & 151.0k & 2.41M \\
    Hungarian & hu & Uralic & Latin & 147.2k & 2.47M \\
    \hline
    Greek & el & Hellenic & Greek & 134.3k & 2.41M \\
    Ukrainian & uk & Slavic & Cyrillic & 108.5k & 1.81M \\
    Indonesian & id & Malayic & Latin & 87.4k & 1.61M \\
    Swedish & sv & Germanic & Latin & 56.6k & 978.0k \\
    \hline
    Total & all & -- & -- & 7.11M & 62.27M \\
	\end{tabular}
	\caption{Size of the \textbf{Top 20 \href{https://github.com/neulab/word-embeddings-for-nmt}{TED Talks}} corpus. English has 253.3k unique lines.}
	\label{tab:ted_training_data}
\end{table*}

\begin{figure*}[b]
    \centering
    \includegraphics[width=0.48\textwidth]{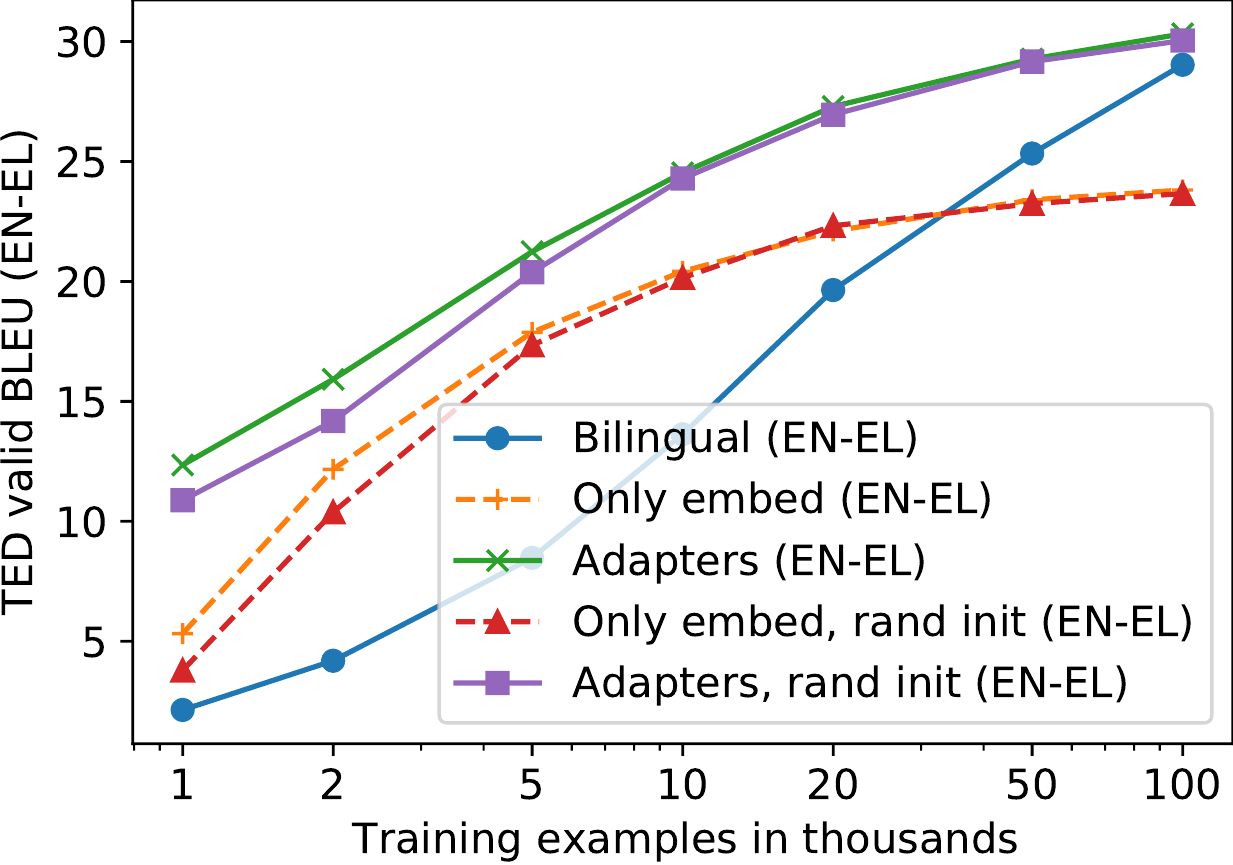}
    \caption{TED validation EN-EL BLEU by training corpus size, with incremental training (\modelref{tgt_embed+lang-code}, \modelref{tgt_embed+adapters+lang-code}) with or without known embedding initialization, versus bilingual baselines \modelrefpar{bilingual}.}
    \label{fig:bleu_by_size_en_el}
\end{figure*}

\newpage

\begin{figure}
    \centering
    \includegraphics[height=5.5cm]{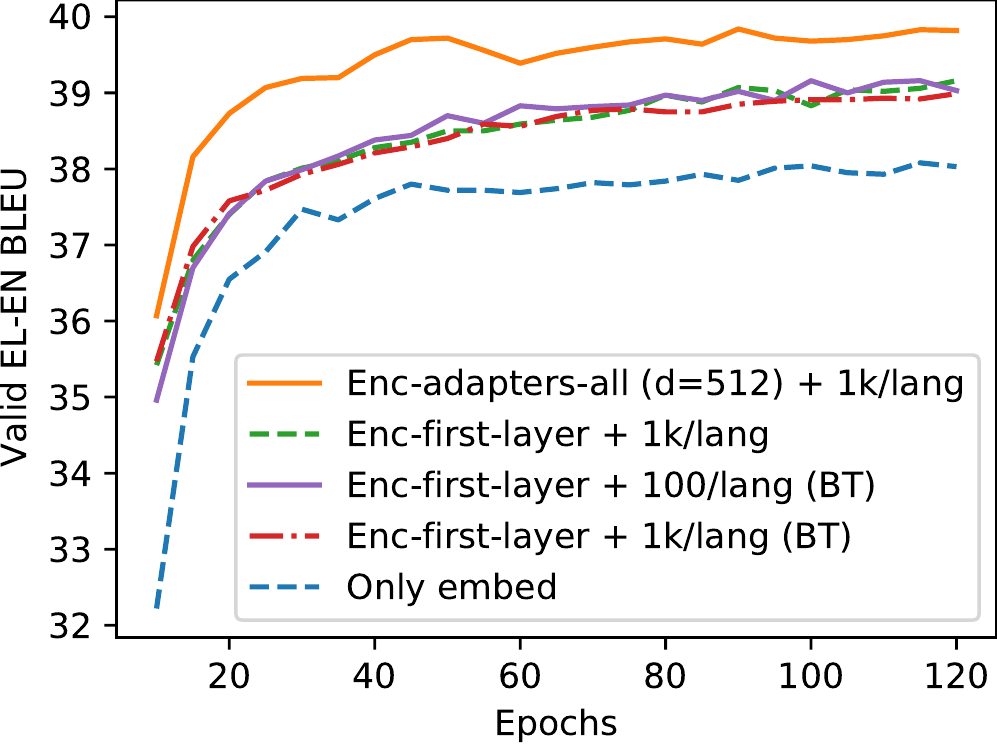}
    \hspace{.1cm}
    \includegraphics[height=5.5cm]{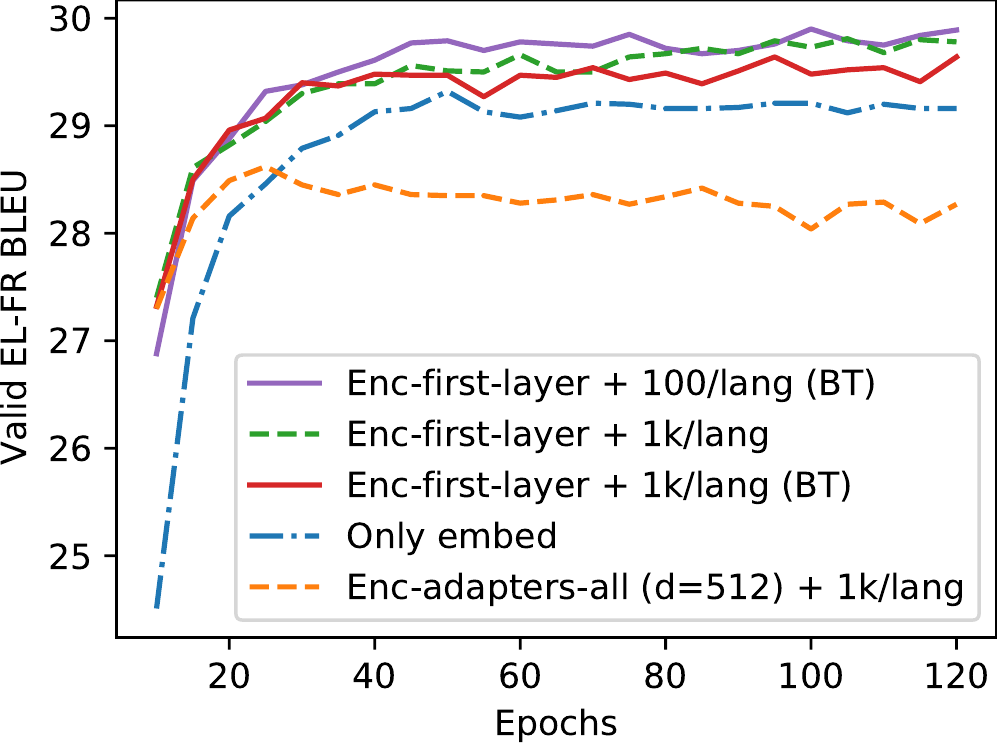}
    \caption{TED validation BLEU on EL-EN (left) and EL-FR (right) while training with EL-EN data, plus some amount of data (real or back-translated) in the 19 other languages.}
    \label{fig:ted_bt}
\end{figure}

\begin{figure}
    \centering
    \includegraphics[height=5.5cm]{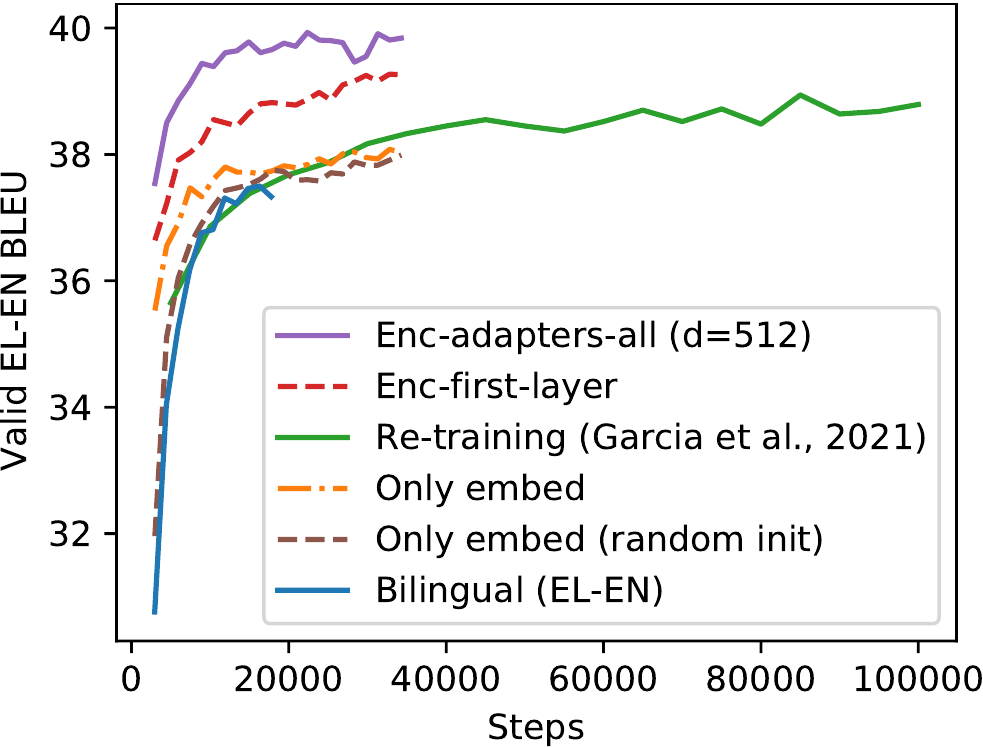}
    \hspace{.1cm}
    \includegraphics[height=5.5cm]{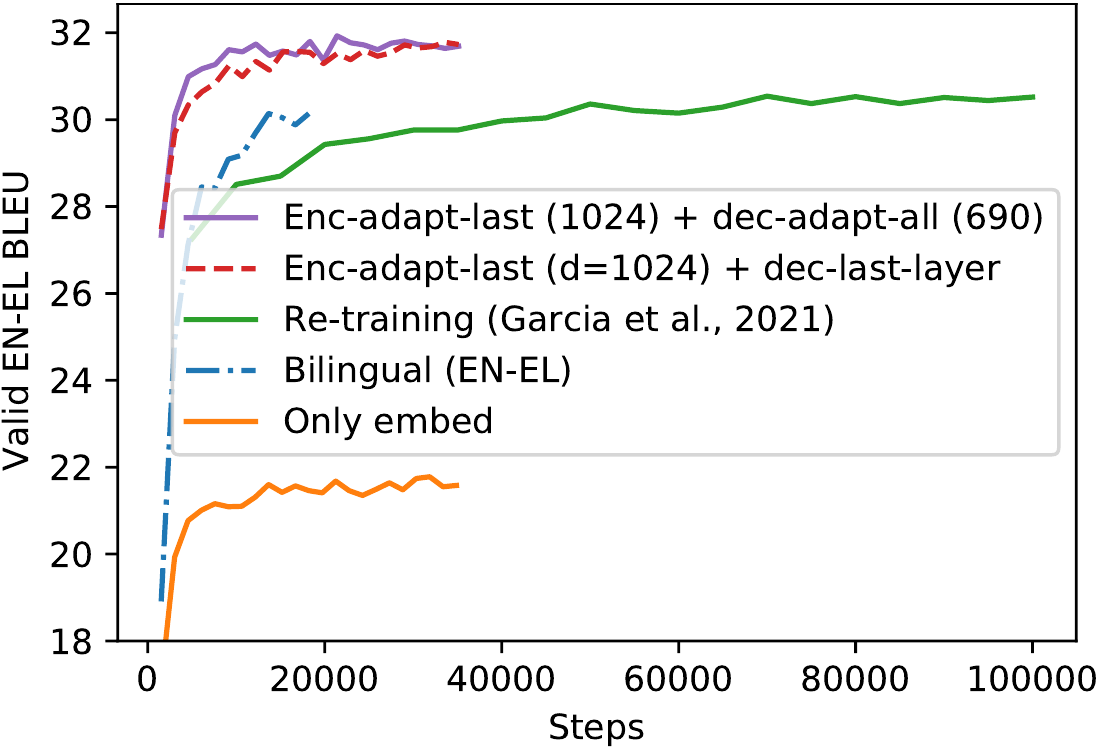}
    \caption{TED validation BLEU on EL-EN (left) and EN-EL (right) while training. Comparison of different incremental training approaches with the \citet{garcia-etal-2021-towards} baseline.}
    \label{fig:ted_train_speed}
\end{figure}

\begin{table*}
    \centering
    \begin{tabular}{c@{\hspace{.2cm}}|c|ccc|ccc}
    \multirow{2}{*}{ID} & \multirow{2}{*}{Model} & \multicolumn{3}{c|}{Valid chrF} & \multicolumn{3}{c}{Test chrF} \\
    & & $\to$EN & $\leftarrow$EN & \nonen{} & $\to$EN & $\leftarrow$EN & \nonen{} \\
    \hline
    \modelref{bilingual} & Bilingual (pivot) & .542 & .484 & .385 & .542 & .484 & .385 \\
    \modelref{en_centric} & English-centric & .530 & .487 & .371 & .529	& .486 & .370 \\
    \modelref{multiparallel} & \modelrefpar{en_centric} + multi-parallel training & .541 & .482 & .395 & .540 & .482 & .395 \\
    \hline
    \setmodel{retraining_en_centric} & \modelrefpar{en_centric} + EL & .528 & \textbf{.488} & .372 & .527 & \textbf{.488} & .371 \\
    \setmodel{retraining_4_en_centric} & \modelrefpar{en_centric} + \{EL, UK, ID, SV\} & .526 & .485 & .370 & .526 & .485 & .370 \\
    \modelref{retraining} & \modelrefpar{multiparallel} + EL & \textbf{.545} & .481 & \textbf{.398} & \textbf{.545} & .482 & \textbf{.398} \\
    \modelref{retraining_4} & \modelrefpar{multiparallel} + \{EL, UK, ID, SV\} & \textbf{.545} & .480 & \textbf{.398} & .544 & .481 & .397 \\
    \hline
    \setmodel{retraining_garcia} & \modelrefpar{multiparallel} + EL \cite{garcia-etal-2021-towards} & .539 & .480 & .394 & .539 & .479 & .394 \\
    \setmodel{retraining_garcia_100k} & \modelrefpar{retraining_garcia} @100k steps & .537 & .479 & .393 & .538 & .478 & .392 \\
    \end{tabular}
    \caption{TED valid and test chrF scores of the baseline models. \modelrefpar{retraining_garcia} corresponds to the best checkpoint according to validation loss (after 3 epochs, or 320k updates) and \modelrefpar{retraining_garcia_100k} is after just 100k updates.}
    \label{tab:ted_baselines_chrf}
\end{table*}

\begin{table*}
    \centering
        \begin{tabular}{c@{\hspace{.2cm}}|c|cc|cc}
        \multirow{2}{*}{ID} & \multirow{2}{*}{Model} & \multicolumn{2}{c|}{Valid chrF} & \multicolumn{2}{c}{Test chrF} \\
        & & \Xtoen{EL} & \Xtononen{EL} & \Xtoen{EL} & \Xtononen{EL} \\
        \hline
        \modelref{bilingual} & Bilingual baselines & .577 & .399 & .583 & .400 \\
        \modelref{retraining} & Re-training + EL & .591 & .426 & .596 & .425 \\
        \modelref{retraining_4} & Re-training + \fourlang{} & .590 & .425 & .594 & .424 \\
        \modelref{retraining_garcia} & Re-training + EL \cite{garcia-etal-2021-towards} & .587 & .424 & .594 & .424 \\
        \modelref{retraining_garcia_100k} & \modelrefpar{retraining_garcia} @100k & .582 & .421 & .587 & .420 \\
        \hline
        \modelref{embed} & Only embed & .577 & .417 & .581 & .417 \\
        \modelref{embed-no-init} & \modelrefpar{embed} + random embed init & .577 & .417$^\star$ & .580 & .417$^\star$ \\
        \modelref{embed+norm+bias} & \modelrefpar{embed} + enc-norm + enc-biases & .582 & .407 & .587 & .407 \\
        \modelref{enc-adapters-1} & \modelrefpar{embed} + enc-adapters-first (d=64) & .578 & .102 & .584 & .100 \\
        \modelref{enc-adapters} & \modelrefpar{embed} + enc-adapters-all (d=64) & .587 & .102 & .593 & .100 \\
        \modelref{enc-adapters-512} & \modelrefpar{embed} + enc-adapters-all (d=512) & .593 & .102 & .602 & .100 \\
        \modelref{enc-adapters-1024-1-2-3} & \modelrefpar{embed} + enc-adapters-\{1,2,3\} (d=1024) & \textbf{.595} & .103 & \textbf{.603} & .101 \\
        \modelref{embed+enc-only-first} & \modelrefpar{embed} + enc-first-layer & .590 & .105 & .595 & .102 \\
        \modelref{only-enc} & \modelrefpar{embed} + enc-all-layers & .590 & .102 & .598 & .100 \\
        \hline
        \modelref{multi_enc-adapters-512} & \modelrefpar{enc-adapters-512} + \multialigned{EL} & .592 & \textbf{.427} & .599 & \textbf{.428} \\
        \modelref{enc-adapters-512+1k} & \modelrefpar{enc-adapters-512} + 1k lines per lang & .594 & .412 & .601 & .413 \\
        \modelref{enc-adapters-512+1k-BT} & \modelrefpar{enc-adapters-512} + 1k lines per lang (BT) & \textbf{.595} & .411 & \textbf{.603} & .411 \\
        \modelref{embed+enc-only-first+1k-BT} & \modelrefpar{embed+enc-only-first} + 1k lines per lang (BT) & .589 & .422 & .596 & .422 \\
        \modelref{enc-adapters-512+100-BT} & \modelrefpar{enc-adapters-512} + 100 lines per lang (BT) & \textbf{.595} & .405 & .601 & .406 \\
        \hline
        \modelref{src4_embed} & \modelrefpar{embed} + \fourlang{} & .582 & .419 & .585 & .419 \\
        \setmodel{src4_enc-adapters-512} & \modelrefpar{enc-adapters-512} + \fourlang{} & .593 & .103 & .597 & .100 \\
        \modelref{src4_embed+enc-only-first} & \modelrefpar{embed+enc-only-first} + \fourlang{} & .587 & .158 & .592 & .154 \\
    \end{tabular}
    \caption{TED valid and test chrF scores of incremental training with Greek on the source side. ($\star$) obtained by using the ``translate into X" lang code embeddings from the initial model. }
    \label{tab:source_el_chrf}
\end{table*}

\begin{table*}
    \centering
        \begin{tabular}{c@{\hspace{.2cm}}|c|cc|cc}
        \multirow{2}{*}{ID} & \multirow{2}{*}{Model} & \multicolumn{2}{c|}{Valid chrF} & \multicolumn{2}{c}{Test chrF} \\
        & & \Xtoen{UK} & \Xtononen{UK} & \Xtoen{UK} & \Xtononen{UK} \\
        \hline
        \modelref{bilingual} & Bilingual baselines & .484 & -- & .494 & -- \\
        \modelref{retraining_4} & Re-training + \fourlang{} & .522 & \textbf{.402} & \textbf{.534} & \textbf{.402} \\
        \hline
        \modelref{embed} & Only embed & .516 & .397 & .525 & .395 \\
        \modelref{embed+norm+bias} & \modelrefpar{embed} + enc-norm + enc-biases & .518 & .386 & .526 & .385 \\
        \modelref{enc-adapters-1} & \modelrefpar{embed} + enc-adapters-first (d=64) & .519 & .100 & .525 & .099 \\
        \modelref{enc-adapters} & \modelrefpar{embed} + enc-adapters-all (d=64) & .520 & .100 & .529 & .100 \\
        \modelref{enc-adapters-512} & \modelrefpar{embed} + enc-adapters-all (d=512) & .520 & .100 & .529 & .099 \\
        \modelref{enc-adapters-1024-1-2-3} & \modelrefpar{embed} + enc-adapters-\{1,2,3\} (d=1024) & .523 & .100 & .530 & .100 \\
        \modelref{embed+enc-only-first} & \modelrefpar{embed} + enc-first-layer & .521 & .136 & .529 & .134 \\
        \modelref{only-enc} & \modelrefpar{embed} + enc-all-layers & .517 & .100 & .525 & .099 \\
        \hline
        \modelref{multi_enc-adapters-512} & \modelrefpar{enc-adapters-512} + \multialigned{UK} & .522 & \textbf{.402} & .530 & .401 \\
        \modelref{enc-adapters-512+1k} & \modelrefpar{enc-adapters-512} + 1k lines per lang & .523 & .389 & .531 & .387 \\
        \modelref{enc-adapters-512+1k-BT} & \modelrefpar{enc-adapters-512} + 1k lines per lang (BT) & .520 & .384 & .528 & .382 \\
        \modelref{embed+enc-only-first+1k-BT} & \modelrefpar{embed+enc-only-first} + 1k lines per lang (BT) & .522 & .397 & .528 & .396 \\
        \modelref{enc-adapters-512+100-BT} & \modelrefpar{enc-adapters-512} + 100 lines per lang (BT) & .520 & .382 & .527 & .382 \\
        \hline
        \modelref{src4_embed} & \modelrefpar{embed} + \fourlang{} & .518 & .398 & .526 & .396 \\
        \modelref{src4_enc-adapters-512} & \modelrefpar{enc-adapters-512} + \fourlang{} & \textbf{.524} & .101 & .532 & .100 \\
        \modelref{src4_embed+enc-only-first} & \modelrefpar{embed+enc-only-first} + \fourlang{} & .522 & .132 & .527 & .130 \\
    \end{tabular}
    \caption{TED valid and test chrF scores of incremental training with Ukrainian on the source side.}
    \label{tab:source_uk_chrf}
\end{table*}

\begin{table*}
    \centering
        \begin{tabular}{c@{\hspace{.2cm}}|c|cc|cc}
        \multirow{2}{*}{ID} & \multirow{2}{*}{Model} & \multicolumn{2}{c|}{Valid chrF} & \multicolumn{2}{c}{Test chrF} \\
        & & \Xtoen{ID} & \Xtononen{ID} & \Xtoen{ID} & \Xtononen{ID} \\
        \hline
        \modelref{bilingual} & Bilingual baselines & .516 & -- & .533 & -- \\
        \modelref{retraining_4} & Re-training + \fourlang{} & .541 & \textbf{.397} & .554 & .404 \\
        \hline
        \modelref{embed} & Only embed & .533 & .390 & .548 & .399 \\
        \modelref{embed-no-init} & \modelrefpar{embed} + random embed init & .529 & .385$^\star$ & .547 & .395$^\star$ \\
        \modelref{embed+norm+bias} & \modelrefpar{embed} + enc-norm + enc-biases & .537 & .379 & .551 & .389 \\
        \modelref{enc-adapters-1} & \modelrefpar{embed} + enc-adapters-first (d=64) & .535 & .100 & .551 & .101 \\
        \modelref{enc-adapters} & \modelrefpar{embed} + enc-adapters-all (d=64) & .540 & .101 & .557 & .102 \\
        \modelref{enc-adapters-512} & \modelrefpar{embed} + enc-adapters-all (d=512) & .541 & .101 & .558 & .102 \\
        \modelref{enc-adapters-1024-1-2-3} & \modelrefpar{embed} + enc-adapters-\{1,2,3\} (d=1024) & \textbf{.545} & .101 & .558 & .102 \\
        \modelref{embed+enc-only-first} & \modelrefpar{embed} + enc-first-layer & .540 & .101 & .554 & .102 \\
        \modelref{only-enc} & \modelrefpar{embed} + enc-all-layers & .535 & .100 & .552 & .101 \\
        \hline
        \modelref{multi_enc-adapters-512} & \modelrefpar{enc-adapters-512} + \multialigned{ID} & .540 & \textbf{.397} & .556 & \textbf{.405} \\
        \modelref{enc-adapters-512+1k} & \modelrefpar{enc-adapters-512} + 1k lines per lang & .541 & .383 & .556 & .390 \\
        \modelref{enc-adapters-512+1k-BT} & \modelrefpar{enc-adapters-512} + 1k lines per lang (BT) & .542 & .382 & .558 & .388 \\
        \modelref{embed+enc-only-first+1k-BT} & \modelrefpar{embed+enc-only-first} + 1k lines per lang (BT) & .536 & .389 & .553 & .399 \\
        \modelref{enc-adapters-512+100-BT} & \modelrefpar{enc-adapters-512} + 100 lines per lang (BT) & .542 & .380 & .557 & .389 \\
        \hline
        \modelref{src4_embed} & \modelrefpar{embed} + \fourlang{} & .530 & .388 & .547 & .397 \\
        \modelref{src4_enc-adapters-512} & \modelrefpar{enc-adapters-512} + \fourlang{} & .543 & .101 & \textbf{.560} & .102 \\
        \modelref{src4_embed+enc-only-first} & \modelrefpar{embed+enc-only-first} + \fourlang{} & .539 & .125 & .553 & .126 \\
    \end{tabular}
    \caption{TED valid and test chrF scores of incremental training with Indonesian on the source side. ($\star$) obtained by using the ``translate into X" lang code embeddings from the initial model. }
    \label{tab:source_id_chrf}
\end{table*}

\begin{table*}
    \centering
        \begin{tabular}{c@{\hspace{.2cm}}|c|cc|cc}
        \multirow{2}{*}{ID} & \multirow{2}{*}{Model} & \multicolumn{2}{c|}{Valid chrF} & \multicolumn{2}{c}{Test chrF} \\
        & & \Xtoen{SV} & \Xtononen{SV} & \Xtoen{SV} & \Xtononen{SV} \\
        \hline
        \modelref{bilingual} & Bilingual baselines & .577 & -- & .579 & -- \\
        \modelref{retraining_4} & Re-training + \fourlang{} & .611 & \textbf{.424} & .615 & \textbf{.426} \\
        \hline
        \modelref{embed} & Only embed & .601 & .417 & .607 & .420 \\
        \modelref{embed-no-init} & \modelrefpar{embed} + random embed init & .596 & .414$^\star$ & .605 & .417$^\star$ \\
        \modelref{embed+norm+bias} & \modelrefpar{embed} + enc-norm + enc-biases & .604 & .413 & .613 & .415 \\
        \modelref{enc-adapters-1} & \modelrefpar{embed} + enc-adapters-first (d=64) & .606 & .100 & .613 & .102 \\
        \modelref{enc-adapters} & \modelrefpar{embed} + enc-adapters-all (d=64) & .610 & .100 & .617 & .102 \\
        \modelref{enc-adapters-512} & \modelrefpar{embed} + enc-adapters-all (d=512) & .608 & .100 & .613 & .102 \\
        \modelref{enc-adapters-1024-1-2-3} & \modelrefpar{embed} + enc-adapters-\{1,2,3\} (d=1024) & .612 & .100 & .619 & .102 \\
        \modelref{embed+enc-only-first} & \modelrefpar{embed} + enc-first-layer & .608 & .102 & .617 & .104 \\
        \modelref{only-enc} & \modelrefpar{embed} + enc-all-layers & .601 & .099 & .605 & .101 \\
        \hline
        \modelref{multi_enc-adapters-512} & \modelrefpar{enc-adapters-512} + \multialigned{SV} & .609 & .421 & .615 & .424 \\
        \modelref{enc-adapters-512+1k} & \modelrefpar{enc-adapters-512} + 1k lines per lang & .606 & .408 & .613 & .410 \\
        \modelref{enc-adapters-512+1k-BT} & \modelrefpar{enc-adapters-512} + 1k lines per lang (BT) & .607 & .406 & .611 & .409 \\
        \modelref{embed+enc-only-first+1k-BT} & \modelrefpar{embed+enc-only-first} + 1k lines per lang (BT) & .605 & .417 & .613 & .420 \\
        \modelref{enc-adapters-512+100-BT} & \modelrefpar{enc-adapters-512} + 100 lines per lang (BT) & .605 & .402 & .611 & .405 \\
        \hline
        \modelref{src4_embed} & \modelrefpar{embed} + \fourlang{} & .601 & .417 & .605 & .418 \\
        \modelref{src4_enc-adapters-512} & \modelrefpar{enc-adapters-512} + \fourlang{} & \textbf{.616} & .100 & \textbf{.620} & .102 \\
        \modelref{src4_embed+enc-only-first} & \modelrefpar{embed+enc-only-first} + \fourlang{} & .607 & .164 & .615 & .172 \\
    \end{tabular}
    \caption{TED valid and test chrF scores of incremental training with Swedish on the source side. ($\star$) obtained by using the ``translate into X" lang code embeddings from the initial model. }
    \label{tab:source_sv_chrf}
\end{table*}

\begin{table*}
    \centering
    \begin{tabular}{c@{\hspace{.2cm}}|c|cc|cc}
        \multirow{2}{*}{ID} & \multirow{2}{*}{Model} & \multicolumn{2}{c|}{Valid chrF} & \multicolumn{2}{c}{Test chrF} \\
        & & \entoX{EL} & \nonentoX{EL} & \entoX{EL} & \nonentoX{EL} \\
        \hline
        \modelref{bilingual} & Bilingual baselines & .551 & .421 & .570 & .432 \\
        \modelref{retraining} & Re-training + EL & .551 & \textbf{.452} & .569 & .460 \\
        \modelref{retraining_4} & Re-training + \fourlang{} & .550 & .451 & .568 & .460 \\
        \modelref{retraining_garcia} & Re-training + EL \cite{garcia-etal-2021-towards} & .553	& .449 & .570 & .458 \\
        \modelref{retraining_garcia_100k} & \modelrefpar{retraining_garcia} @100k & .553 & .450 & .572 & .459 \\
        \hline
        \modelref{tgt_embed+lang-code} & Only embed & .504 & .415 & .518 & .423 \\
        \modelref{tgt_embed+non-tied+lang-code} & \modelrefpar{tgt_embed+lang-code} + non-tied & .517 & .423 & .530 & .432 \\
        \modelref{tgt_embed+dec-adapters+lang-code} & \modelrefpar{tgt_embed+lang-code} + dec-adapters-all (d=64) & .533 & .435 & .551 & .444 \\
        \modelref{tgt_embed+adapters+lang-code} & \modelrefpar{tgt_embed+lang-code} + adapters-all (d=64) & .556 & .440 & .574 & .449 \\
        \modelref{tgt_embed+enc-adapters-6-1024+lang-code} & \modelrefpar{tgt_embed+lang-code} + enc-adapters-last (d=1024) & .555 & .445 & .571 & .455 \\
        \modelref{tgt_embed+enc-adapters-6-1024+dec-adapters+lang-code} & \modelrefpar{tgt_embed+dec-adapters+lang-code} + enc-adapters-last (d=1024) & .560 & .450 & .580 & .459 \\
        \modelref{tgt_embed+dec-only-last+lang-code} & \modelrefpar{tgt_embed+lang-code} + dec-last-layer & .556 & .449 & .576 & .458 \\
        \modelref{tgt_embed+dec-only-last+enc-adapters-6-1024+lang-code} & \modelrefpar{tgt_embed+dec-only-last+lang-code} + enc-adapters-last (d=1024) & .566 & \textbf{.452} & .585 & .461 \\
        \modelref{tgt_embed+adapters-430+lang-code} & \modelrefpar{tgt_embed+lang-code} + adapters-all (d=430) & .566 & .426 & .585 & .436 \\
        \modelref{tgt_embed+enc-adapters-6-1024+dec-adapters-690+lang-code} & \modelrefpar{tgt_embed+lang-code} + dec-ad-all (690) + enc-ad-last (1024) & .566 & .451 & .584 & .461 \\
        \modelref{tgt_embed+dec-only-last+adapters-90+lang-code} & \modelrefpar{tgt_embed+dec-only-last+lang-code} + adapters-all (d=90) & .565 & .442 & .585 & .453 \\
        \modelref{tgt_embed+dec-only-last+enc-adapters-170+lang-code} & \modelrefpar{tgt_embed+dec-only-last+lang-code} + enc-adapters-all (d=170) & \textbf{.567} & .435 & \textbf{.586} & .444 \\
        \setmodel{tgt_embed+new-dec-6+lang-code} & \modelrefpar{tgt_embed+lang-code} + dec-all-layers & .562 & .448 & .577 & .457 \\
        \hline
        \modelref{tgt_multi_embed+dec-only-last+enc-adapters-6-1024+lang-code} & \modelrefpar{tgt_embed+dec-only-last+enc-adapters-6-1024+lang-code} + \multialigned{EL} & .554 & .451 & .573 & \textbf{.462} \\
        \hline
        \modelref{tgt_inc4_embed+dec-only-last+enc-adapters-6-1024+lang-code} & \modelrefpar{tgt_embed+dec-only-last+enc-adapters-6-1024+lang-code} + \fourlang{} & .558 & .449 & .578 & .458 \\
    \end{tabular}
    \caption{TED valid and test chrF scores of incremental training with Greek on the target side.}
    \label{tab:target_el_chrf}
\end{table*}

\begin{table*}
    \centering
    \begin{tabular}{c@{\hspace{.2cm}}|c|cc|cc}
        \multirow{2}{*}{ID} & \multirow{2}{*}{Model} & \multicolumn{2}{c|}{Valid chrF} & \multicolumn{2}{c}{Test chrF} \\
        & & \entoX{UK} & \nonentoX{UK} & \entoX{UK} & \nonentoX{UK} \\
        \hline
        \modelref{bilingual} & Bilingual baselines & .441 & -- & .440 & -- \\
        \modelref{retraining_4} & Re-training + \fourlang{} & .460 & .401 & .459 & .394 \\
        \hline
        \modelref{tgt_embed+lang-code} & Only embed & .446 & .386 & .445 & .380 \\
        \modelref{tgt_embed+non-tied+lang-code} & \modelrefpar{tgt_embed+lang-code} + non-tied & .452 & .390 & .449 & .384 \\
        \modelref{tgt_embed+dec-adapters+lang-code} & \modelrefpar{tgt_embed+lang-code} + dec-adapters-all (d=64) & .457 & .394 & .453 & .387 \\
        \modelref{tgt_embed+adapters+lang-code} & \modelrefpar{tgt_embed+lang-code} + adapters-all (d=64) & .466 & .397 & .465 & .391 \\
        \modelref{tgt_embed+enc-adapters-6-1024+lang-code} & \modelrefpar{tgt_embed+lang-code} + enc-adapters-last (d=1024) & .464 & .400 & .463 & .393 \\
        \modelref{tgt_embed+enc-adapters-6-1024+dec-adapters+lang-code} & \modelrefpar{tgt_embed+dec-adapters+lang-code} + enc-adapters-last (d=1024) & .469 & .402 & .465 & .394 \\
        \modelref{tgt_embed+dec-only-last+lang-code} & \modelrefpar{tgt_embed+lang-code} + dec-last-layer & .468 & .402 & .464 & .394 \\
        \modelref{tgt_embed+dec-only-last+enc-adapters-6-1024+lang-code} & \modelrefpar{tgt_embed+dec-only-last+lang-code} + enc-adapters-last (d=1024) & \textbf{.471} & \textbf{.403} & .467 & \textbf{.395} \\
        \modelref{tgt_embed+adapters-430+lang-code} & \modelrefpar{tgt_embed+lang-code} + adapters-all (d=430) & .470 & .386 & .468 & .380 \\
        \modelref{tgt_embed+enc-adapters-6-1024+dec-adapters-690+lang-code} & \modelrefpar{tgt_embed+lang-code} + dec-ad-all (690) + enc-ad-last (1024) & .468 & .402 & .466 & .394 \\
        \modelref{tgt_embed+dec-only-last+adapters-90+lang-code} & \modelrefpar{tgt_embed+dec-only-last+lang-code} + adapters-all (d=90) & .469 & .400 & .467 & .393 \\
        \modelref{tgt_embed+dec-only-last+enc-adapters-170+lang-code} & \modelrefpar{tgt_embed+dec-only-last+lang-code} + enc-adapters-all (d=170) & .470 & .393 & \textbf{.469} & .387 \\
        \modelref{tgt_embed+new-dec-6+lang-code} & \modelrefpar{tgt_embed+lang-code} + dec-all-layers & .468 & .400 & .463 & .391 \\
        \hline
        \modelref{tgt_multi_embed+dec-only-last+enc-adapters-6-1024+lang-code} & \modelrefpar{tgt_embed+dec-only-last+enc-adapters-6-1024+lang-code} + \multialigned{UK} & .460 & .400 & .458 & .393 \\
        \hline
        \modelref{tgt_inc4_embed+dec-only-last+enc-adapters-6-1024+lang-code} & \modelrefpar{tgt_embed+dec-only-last+enc-adapters-6-1024+lang-code} + \fourlang{} & .467 & .402 & .465 & \textbf{.395} \\
    \end{tabular}
    \caption{TED valid and test chrF scores of incremental training with Ukrainian on the target side.}
    \label{tab:target_uk_chrf}
\end{table*}

\begin{table*}
    \centering
    \begin{tabular}{c@{\hspace{.2cm}}|c|cc|cc}
        \multirow{2}{*}{ID} & \multirow{2}{*}{Model} & \multicolumn{2}{c|}{Valid chrF} & \multicolumn{2}{c}{Test chrF} \\
        & & \entoX{ID} & \nonentoX{ID} & \entoX{ID} & \nonentoX{ID} \\
        \hline
        \modelref{bilingual} & Bilingual baselines & .568 & -- & .579 & -- \\
        \modelref{retraining_4} & Re-training + \fourlang{} & .579 & \textbf{.498} & .591 & .504 \\
        \hline
        \modelref{tgt_embed+lang-code} & Only embed & .562 & .483 & .575 & .491 \\
        \modelref{tgt_embed+non-tied+lang-code} & \modelrefpar{tgt_embed+lang-code} + non-tied & .569 & .487 & .582 & .496 \\
        \modelref{tgt_embed+dec-adapters+lang-code} & \modelrefpar{tgt_embed+lang-code} + dec-adapters-all (d=64) & .579 & .492 & .591 & .501 \\
        \modelref{tgt_embed+adapters+lang-code} & \modelrefpar{tgt_embed+lang-code} + adapters-all (d=64) & .585 & .489 & .599 & .498 \\
        \modelref{tgt_embed+enc-adapters-6-1024+lang-code} & \modelrefpar{tgt_embed+lang-code} + enc-adapters-last (d=1024) & .586 & .493 & .600 & .501 \\
        \modelref{tgt_embed+enc-adapters-6-1024+dec-adapters+lang-code} & \modelrefpar{tgt_embed+dec-adapters+lang-code} + enc-adapters-last (d=1024) & \textbf{.589} & .495 & .602 & .504 \\
        \modelref{tgt_embed+dec-only-last+lang-code} & \modelrefpar{tgt_embed+lang-code} + dec-last-layer & .588 & .496 & .598 & .503 \\
        \modelref{tgt_embed+dec-only-last+enc-adapters-6-1024+lang-code} & \modelrefpar{tgt_embed+dec-only-last+lang-code} + enc-adapters-last (d=1024) & .587 & .496 & .601 & .503 \\
        \modelref{tgt_embed+adapters-430+lang-code} & \modelrefpar{tgt_embed+lang-code} + adapters-all (d=430) & \textbf{.589} & .480 & .599 & .489 \\
        \modelref{tgt_embed+enc-adapters-6-1024+dec-adapters-690+lang-code} & \modelrefpar{tgt_embed+lang-code} + dec-ad-all (690) + enc-ad-last (1024) & .586 & .494 & .599 & .502 \\
        \modelref{tgt_embed+dec-only-last+adapters-90+lang-code} & \modelrefpar{tgt_embed+dec-only-last+lang-code} + adapters-all (d=90) & .588 & .493 & \textbf{.603} & .502 \\
        \modelref{tgt_embed+dec-only-last+enc-adapters-170+lang-code} & \modelrefpar{tgt_embed+dec-only-last+lang-code} + enc-adapters-all (d=170) & \textbf{.589} & .489 & .602 & .496 \\
        \modelref{tgt_embed+new-dec-6+lang-code} & \modelrefpar{tgt_embed+lang-code} + dec-all-layers & .588 & .497 & .600 & \textbf{.506} \\
        \hline
        \modelref{tgt_multi_embed+dec-only-last+enc-adapters-6-1024+lang-code} & \modelrefpar{tgt_embed+dec-only-last+enc-adapters-6-1024+lang-code} + \multialigned{ID} & .579 & .496 & .589 & .502 \\
        \hline
        \modelref{tgt_inc4_embed+dec-only-last+enc-adapters-6-1024+lang-code} & \modelrefpar{tgt_embed+dec-only-last+enc-adapters-6-1024+lang-code} + \fourlang{} & .584 & .493 & .597 & .501 \\
    \end{tabular}
    \caption{TED valid and test chrF scores of incremental training with Indonesian on the target side.}
    \label{tab:target_id_chrf}
\end{table*}

\begin{table*}
    \centering
    \begin{tabular}{c@{\hspace{.2cm}}|c|cc|cc}
        \multirow{2}{*}{ID} & \multirow{2}{*}{Model} & \multicolumn{2}{c|}{Valid chrF} & \multicolumn{2}{c}{Test chrF} \\
        & & \entoX{SV} & \nonentoX{SV} & \entoX{SV} & \nonentoX{SV} \\
        \hline
        \modelref{bilingual} & Bilingual baselines & .557 & -- & .557 & -- \\
        \modelref{retraining_4} & Re-training + \fourlang{} & .568 & .453 & .567 & .455 \\
        \hline
        \modelref{tgt_embed+lang-code} & Only embed & .547 & .436 & .548 & .438 \\
        \modelref{tgt_embed+non-tied+lang-code} & \modelrefpar{tgt_embed+lang-code} + non-tied & .552 & .441 & .556 & .443 \\
        \modelref{tgt_embed+dec-adapters+lang-code} & \modelrefpar{tgt_embed+lang-code} + dec-adapters-all (d=64) & .569 & .451 & .572 & .453 \\
        \modelref{tgt_embed+adapters+lang-code} & \modelrefpar{tgt_embed+lang-code} + adapters-all (d=64) & .584 & .444 & .589 & .448 \\
        \modelref{tgt_embed+enc-adapters-6-1024+lang-code} & \modelrefpar{tgt_embed+lang-code} + enc-adapters-last (d=1024) & .583 & .451 & .586 & .455 \\
        \modelref{tgt_embed+enc-adapters-6-1024+dec-adapters+lang-code} & \modelrefpar{tgt_embed+dec-adapters+lang-code} + enc-adapters-last (d=1024) & .587 & .452 & .590 & .457 \\
        \modelref{tgt_embed+dec-only-last+lang-code} & \modelrefpar{tgt_embed+lang-code} + dec-last-layer & .580 & .452 & .584 & .458 \\
        \modelref{tgt_embed+dec-only-last+enc-adapters-6-1024+lang-code} & \modelrefpar{tgt_embed+dec-only-last+lang-code} + enc-adapters-last (d=1024) & \textbf{.589} & .454 & .590 & .458 \\
        \modelref{tgt_embed+adapters-430+lang-code} & \modelrefpar{tgt_embed+lang-code} + adapters-all (d=430) & .587 & .437 & .597 & .443 \\
        \modelref{tgt_embed+enc-adapters-6-1024+dec-adapters-690+lang-code} & \modelrefpar{tgt_embed+lang-code} + dec-ad-all (690) + enc-ad-last (1024) & .584 & .452 & .587 & .457 \\
        \modelref{tgt_embed+dec-only-last+adapters-90+lang-code} & \modelrefpar{tgt_embed+dec-only-last+lang-code} + adapters-all (d=90) & .587 & .449 & .590 & .453 \\
        \modelref{tgt_embed+dec-only-last+enc-adapters-170+lang-code} & \modelrefpar{tgt_embed+dec-only-last+lang-code} + enc-adapters-all (d=170) & .588 & .443 & \textbf{.591} & .449 \\
        \modelref{tgt_embed+new-dec-6+lang-code} & \modelrefpar{tgt_embed+lang-code} + dec-all-layers & .583 & .452 & .586 & .457 \\
        \hline
        \modelref{tgt_multi_embed+dec-only-last+enc-adapters-6-1024+lang-code} & \modelrefpar{tgt_embed+dec-only-last+enc-adapters-6-1024+lang-code} + \multialigned{SV} & .570 & .454 & .569 & .456 \\
        \hline
        \modelref{tgt_inc4_embed+dec-only-last+enc-adapters-6-1024+lang-code} & \modelrefpar{tgt_embed+dec-only-last+enc-adapters-6-1024+lang-code} + \fourlang{} & \textbf{.589} & \textbf{.455} & \textbf{.591} & \textbf{.459} \\
    \end{tabular}
    \caption{TED valid and test chrF scores of incremental training with Swedish on the target side.}
    \label{tab:target_sv_chrf}
\end{table*}

\begin{table*}
    \centering
        \begin{tabular}{c@{\hspace{.2cm}}|c|cc|cc}
        \multirow{2}{*}{ID} & \multirow{2}{*}{Model} & \multicolumn{2}{c|}{Valid chrF} & \multicolumn{2}{c}{Test chrF} \\
        & & \Xtoen{EL} & \Xtononen{EL} & \Xtoen{EL} & \Xtononen{EL} \\
        \hline
        \modelref{bilingual} & Bilingual baselines (pivot) & .577 & .407 & .583 & .407 \\
        \modelref{retraining_en_centric} & Re-training + EL (pivot) & .575	& \textbf{.408} & .578 & \textbf{.408} \\
        \modelref{retraining_4_en_centric} & Re-training + \fourlang{} (pivot) & .572	& .405 & .573 & .404 \\
        \hline
        \modelref{embed} & Only embed & .571 & .402 & .579 & .403 \\
        \modelref{embed+norm+bias} & \modelrefpar{embed} + enc-norm + enc-biases & .574 & .403 & .581 & .403 \\
        \modelref{enc-adapters-1} & \modelrefpar{embed} + enc-adapters-first (d=64) & .575 & .404 & .582 & .404 \\
        \modelref{enc-adapters} & \modelrefpar{embed} + enc-adapters-all (d=64) & .581 & .401 & .587 & .401 \\
        \modelref{enc-adapters-512} & \modelrefpar{embed} + enc-adapters-all (d=512) & \textbf{.584} & .386 & .589 & .387 \\
        \modelref{enc-adapters-1024-1-2-3} & \modelrefpar{embed} + enc-adapters-\{1,2,3\} (d=1024) & \textbf{.584} & .403 & \textbf{.591} & .404 \\
        \modelref{embed+enc-only-first} & \modelrefpar{embed} + enc-first-layer & .581 & .114 & .588 & .111 \\
        \modelref{only-enc} & \modelrefpar{embed} + enc-all-layers & .579 & .102 & .590 & .100 \\
        \hline
        \modelref{enc-adapters-512+1k} & \modelrefpar{enc-adapters-512} + 1k lines per lang & .581 & .394 & .589 & .393 \\
        \modelref{enc-adapters-512+1k-BT} & \modelrefpar{enc-adapters-512} + 1k lines per lang (BT) & .583 & .396 & .588 & .396 \\
        \modelref{embed+enc-only-first+1k-BT} & \modelrefpar{embed+enc-only-first} + 1k lines per lang (BT) & .579 & .404 & .587 & .404 \\
        \modelref{enc-adapters-512+100-BT} & \modelrefpar{enc-adapters-512} + 100 lines per lang (BT) & .583 & .391 & .590 & .391 \\
    \end{tabular}
    \caption{TED valid and test chrF scores of incremental training with Greek on the source side when the initial model is English-centric (\modelref{en_centric}).}
    \label{tab:source_greek_en}
\end{table*}

\begin{table*}
    \centering
        \begin{tabular}{c@{\hspace{.2cm}}|c|cc|cc}
        \multirow{2}{*}{ID} & \multirow{2}{*}{Model} & \multicolumn{2}{c|}{Valid chrF} & \multicolumn{2}{c}{Test chrF} \\
        & & \entoX{EL} & \nonentoX{EL} & \entoX{EL} & \nonentoX{EL} \\
        \hline
        \modelref{bilingual} & Bilingual baselines (pivot) & .551 & .435 & .570 & .444 \\
        \modelref{retraining_en_centric} & Re-training + EL (pivot) & .557 & .430 & .577 & .440 \\
        \modelref{retraining_4_en_centric} & Re-training + \fourlang{} (pivot) & .556	& .428 & .573 & .438 \\
        \hline
        \modelref{tgt_embed+lang-code} & Only embed & .518 & .401 & .534 & .409 \\
        \modelref{tgt_embed+non-tied+lang-code} & \modelrefpar{tgt_embed+lang-code} + non-tied & .529 & .407 & .545 & .416 \\
        \modelref{tgt_embed+dec-adapters+lang-code} & \modelrefpar{tgt_embed+lang-code} + dec-adapters-all (d=64) & .545 & .415 & .561 & .424 \\
        \modelref{tgt_embed+adapters+lang-code} & \modelrefpar{tgt_embed+lang-code} + adapters-all (d=64) & .564 & .414 & .584 & .423 \\
        \modelref{tgt_embed+enc-adapters-6-1024+lang-code} & \modelrefpar{tgt_embed+lang-code} + enc-adapters-last (d=1024) & .562 & .420 & .581 & .429 \\
        \modelref{tgt_embed+enc-adapters-6-1024+dec-adapters+lang-code} & \modelrefpar{tgt_embed+dec-adapters+lang-code} + enc-adapters-last (d=1024) & .568 & .424 & .586 & .432 \\
        \modelref{tgt_embed+dec-only-last+lang-code} & \modelrefpar{tgt_embed+lang-code} + dec-last-layer & .565 & .427 & .585 & .435 \\
        \modelref{tgt_embed+dec-only-last+enc-adapters-6-1024+lang-code} & \modelrefpar{tgt_embed+dec-only-last+lang-code} + enc-adapters-last (d=1024) & .571 & .426 & .589 & .435 \\
        \modelref{tgt_embed+adapters-430+lang-code} & \modelrefpar{tgt_embed+lang-code} + adapters-all (d=430) & \textbf{.573} & .408 & .591 & .416 \\
        \modelref{tgt_embed+enc-adapters-6-1024+dec-adapters-690+lang-code} & \modelrefpar{tgt_embed+lang-code} + dec-ad-all (690) + enc-ad-last (1024) & .568 & .425 & .589 & .434 \\
        \modelref{tgt_embed+dec-only-last+adapters-90+lang-code} & \modelrefpar{tgt_embed+dec-only-last+lang-code} + adapters-all (d=90) & \textbf{.573} & .420 & .592 & .428 \\
        \modelref{tgt_embed+dec-only-last+enc-adapters-170+lang-code} & \modelrefpar{tgt_embed+dec-only-last+lang-code} + enc-adapters-all (d=170) & .572 & .417 & \textbf{.594} & .426 \\
        \modelref{tgt_embed+new-dec-6+lang-code} & \modelrefpar{tgt_embed+lang-code} + dec-all-layers & .566 & .426 & .583 & .434 \\
        \hline
        \modelref{tgt_multi_embed+dec-only-last+enc-adapters-6-1024+lang-code} & \modelrefpar{tgt_embed+dec-only-last+enc-adapters-6-1024+lang-code} + \multialigned{EL} & .553 & \textbf{.441} & .570 & \textbf{.450} \\
    \end{tabular}
    \caption{TED valid and test chrF scores of incremental training with Greek on the target side when the initial model is English-centric (\modelref{en_centric}).}
    \label{tab:target_greek_en}
\end{table*}

\begin{table*}
    \centering
    \begin{tabular}{cc|cc|c|c}
        & Source model & & Target model & Valid & Test \\
        \hline
        \modelref{bilingual} & \multicolumn{3}{c|}{Bilingual} & .384 & .388 \\
        \modelref{bilingual} & \multicolumn{3}{c|}{Bilingual (pivot through English)} & .428 & .437 \\
        \modelref{retraining_4} & \multicolumn{3}{c|}{Re-training + \fourlang{}} & \textbf{.461} & \textbf{.470} \\
        \hline
        \multirow{2}{*}{\modelref{embed}} & \multirow{2}{*}{Only embed} & \modelref{tgt_embed+dec-only-last+lang-code} & Dec-last-layer & .456 & .465 \\
        & & \modelref{tgt_embed+dec-only-last+enc-adapters-6-1024+lang-code} & Dec-last-layer + enc-ad-last (d=1024) & .457 & .466 \\
        \hline
        \multirow{3}{*}{\modelref{embed+enc-only-first}} & \multirow{3}{*}{Enc-first-layer} & \modelref{tgt_embed+dec-only-last+lang-code} & Dec-last-layer & .453 & .463 \\
        & & \modelref{tgt_embed+dec-only-last+enc-adapters-6-1024+lang-code} & Dec-last-layer + enc-ad-last (d=1024) & \textbf{.461} & \textbf{.470} \\
        & & \modelref{tgt_embed+dec-only-last+enc-adapters-6-1024+lang-code} & Pivot through English$^\star$ & .460 & .469 \\
        \hline
        \multirow{2}{*}{\modelref{embed+enc-only-first+1k-BT}} & \multirow{2}{*}{Enc-first-layer + 1k (BT)} & \modelref{tgt_embed+dec-only-last+lang-code} & Dec-last-layer & .459 & .468 \\
        & & \modelref{tgt_embed+dec-only-last+enc-adapters-6-1024+lang-code} & Dec-last-layer + enc-ad-last (d=1024) & \textbf{.461} & \textbf{.470} \\
        \hline
        \multirow{2}{*}{\modelref{enc-adapters-512+1k-BT}} &  \multirow{2}{*}{Enc-adapters-all (d=512) + 1k (BT)} & \modelref{tgt_embed+dec-only-last+lang-code} & Dec-last-layer & .451 & .459 \\
        & & \modelref{tgt_embed+dec-only-last+enc-adapters-6-1024+lang-code} & Dec-last-layer + enc-ad-last (d=1024) & .453 & .462 \\
    \end{tabular}
    \caption{TED valid and test chrF scores on \fourlang{}$\to$\fourlang{} (average over 12 directions) by combining source-language and target-language incrementally-trained parameters. ($\star$) instead of combining model parameters, translate with \modelrefpar{embed+enc-only-first} to English, then to the target language with \modelrefpar{tgt_embed+dec-only-last+enc-adapters-6-1024+lang-code}.}
    \label{tab:source_target_inc_chrf}
\end{table*}

\begin{table}[h]
	\centering
	\begin{tabular}{c|c|c|cH|cHH}
    Language & Code & Family & X-EN lines & Trg. prob (T=5) & X-* lines & Trg. prob (T=5) & Trg. prob (T=2) \\
    \hline
    English & en & Germanic & 450.30M & 0.500 & 450.30M & 0.068 & 0.109 \\
    French & fr & Romance & 95.43M & 0.038 & 215.63M & 0.056 & 0.070 \\
    German & de & Romance & 76.49M & 0.036 & 192.67M & 0.056 & 0.068 \\
    Spanish & es & Romance & 72.97M & 0.036 & 191.71M & 0.056 & 0.068 \\
    Italian & it & Romance & 38.05M & 0.031 & 136.11M & 0.054 & 0.061 \\
    Portuguese & pt & Romance & 29.18M & 0.030 & 117.68M & 0.053 & 0.058 \\
    Dutch & nl & Germanic & 27.36M & 0.029 & 104.35M & 0.052 & 0.055 \\
    Norwegian & nb & Germanic & 15.38M & 0.026 & 65.37M & 0.049 & 0.045 \\
    Czech & cs & Slavic & 12.92M & 0.025 & 65.55M & 0.049 & 0.046 \\
    Polish & pl & Slavic & 12.88M & 0.025 & 69.27M & 0.050 & 0.047 \\
    Swedish & sv & Germanic & 10.97M & 0.025 & 60.16M & 0.048 & 0.044 \\
    Danish & da & Germanic & 9.79M & 0.024 & 61.28M & 0.049 & 0.044 \\
    Greek$^\star$ & el & Hellenic & 8.92M & 0.024 & 48.29M & 0.046 & 0.039 \\
    Finnish & fi & Uralic & 6.83M & 0.022 & 47.62M & 0.047 & 0.040 \\
    Croatian & hr & Slavic & 6.34M & 0.022 & 30.47M & 0.042 & 0.031 \\
    Hungarian & hu & Uralic & 6.29M & 0.022 & 42.53M & 0.046 & 0.037 \\
    Bulgarian$^\star$ & bg & Slavic & 6.10M & 0.022 & 36.84M & 0.044 & 0.035 \\
    Romanian & ro & Romance & 5.79M & 0.022 & 40.52M & 0.045 & 0.037 \\
    Slovak & sk & Slavic & 4.56M & 0.021 & 36.39M & 0.044 & 0.035 \\
    Lithuanian & lt & Baltic & 4.03M & 0.020 & 30.21M & 0.043 & 0.032 \\
    \hline
    Total & all & -- & 900.60M & 1.0 & 2.043B & 1.0 & 1.0 \\
	\end{tabular}
	\caption{Size of the \textbf{Top 20 \href{https://paracrawl.eu/v7-1}{ParaCrawl}} corpus. English has 271.85M unique lines. ($\star$) all languages use the Latin script, except for Greek and Bulgarian (Cyrillic).}
	\label{tab:paracrawl_training_data}
\end{table}

\begin{table}[h]
	\centering
	\begin{tabular}{c|c|c|cHHH}
    Language & Code & Family & X-EN lines & X-RU lines & X-AR lines & X-ZH lines \\
    \hline
    Russian & ru & Slavic & 25.17M & -- & 14.03M & 12.33M \\
    Arabic & ar & Semitic & 20.04M & 14.03M & -- & 11.91M \\
    Mandarin Chinese & zh & Sinitic & 17.45M & 12.33M & 11.91M & -- \\
    \end{tabular}
    \caption{Size of the UNPC corpus.}
    \label{tab:unpc_training_data}
\end{table}

\begin{table}
	\centering
 	\small
	\begin{tabular}{|cc|}
		\hline
		Parameter name & Parameter value \\
		\hline
		share\_all\_embeddings & True / False$^{5,6}$ \\
		share\_decoder\_input\_output\_embed & True \\
		arch & transformer \\
		lr\_scheduler & inverse\_sqrt \\
		optimizer & adam \\
		adam\_betas & 0.9,0.999 \\
		fp16 & True \\
		clip\_norm & 0.0 \\
		lr & 0.0005 / 0.0001$^{4}$ \\
		warmup\_updates & 4000 \\
		warmup\_init\_lr & 1e-07 \\
		criterion & label\_smoothed\_cross\_entropy \\
		label\_smoothing & 0.1 \\
		dropout & 0.3 / 0.1$^{2,3,4}$ \\
		max\_tokens & 4000 \\
		max\_epoch & 120$^{1,5}$ / 10$^{2,4}$ / 20$^{3,6}$ \\
		save-interval & 1 / 5$^{5}$ \\
		validate-interval & 1 / 5$^{5}$ \\
		update\_freq$^\star$ & 4 \\
		reset\_* & True \\
		\hline
		lang\_temperature$^\dagger$ & 5 \\
		\hline
	\end{tabular}
	\caption{\small fairseq v0.10.2 hyper-parameters of the \textbf{TED Talks models}. ($\star$) we normalize this value by the number of GPUs to have a constant batch size. For instance, models trained on 4 GPUs use \texttt{update\_freq=1}. (1) English-centric training stage of the initial model; (2) multi-parallel training stage; (3) our re-training approach; (4) our implementation of \citet{garcia-etal-2021-towards}; (5) English-centric incremental training; (6) multi-aligned incremental training.  The bilingual baselines use the \texttt{transformer\_iwslt\_de\_en} architecture and are trained for 25k steps with validation every 500 steps and patience 3. ($\dagger$) we implement an on-the-fly data loading pipeline that builds heterogeneous batches by sampling language pair $k$ with probability: $p_k=D^{1/T}_k/(\sum{D^{1/T}_i})$ where $T$ is the temperature and $D_k$ is the total number of line pairs for that language pair \cite{aharoni-etal-2019-massively}.}
	\label{tab:ted_hyperparameters}
\end{table}

\begin{table}
	\centering
 	\small
	\begin{tabular}{|cc|}
		\hline
		Parameter name & Parameter value \\
		\hline
	    max\_source\_positions & 256 \\
		max\_target\_positions & 256 \\
		share\_all\_embeddings & True / False$^{5}$ \\
		share\_decoder\_input\_output\_embed & True \\
		arch & transformer\_vaswani\_wmt\_en\_de\_big \\
		encoder\_layers & 12 / 6$^{4}$ \\
		decoder\_layers & 2 / 6$^{4}$ \\
		lr\_scheduler & inverse\_sqrt \\
		optimizer & adam \\
		adam\_betas & 0.9,0.98 \\
		fp16 & True \\
		clip\_norm & 1.0 \\
		lr & 0.0005 \\
		warmup\_updates & 4000 \\
		warmup\_init\_lr & 1e-07 \\
		criterion & label\_smoothed\_cross\_entropy \\
		label\_smoothing & 0.1 \\
		dropout & 0.1 \\
		max\_tokens & 8000 \\
		max\_update & 1000000$^1$ / 200000$^2$ / 360000$^3$ / 120000$^{4,5}$ \\
		save\_interval\_updates & 20000 / 10000$^4$ / 5000$^5$ \\
		validate\_interval\_updates & 20000 / 10000$^4$ / 5000$^5$ \\
		update\_freq$^\star$ & 32 \\
		reset\_* & True \\
		\hline
		lang\_temperature$^\dagger$ & 5 / 2$^{2,3}$ \\
		\hline
	\end{tabular}
	\caption{\small fairseq hyper-parameters of the \textbf{ParaCrawl/UNPC models}. ($\star$) ($\dagger$) see Table~\ref{tab:ted_hyperparameters}. (1) English-centric training stage of the initial model; (2) multi-parallel training stage; (3) our re-training approach; (4) bilingual baselines; (5) incremental training.}
	\label{tab:paracrawl_hyperparameters}
\end{table}

\begin{table}[]
    \centering
    \begin{tabular}{c@{\hspace{.2cm}}|c|c|cccc|cccc}
        \multirow{2}{*}{ID} & \multirow{2}{*}{Model} & \multirow{2}{*}{Lang code} & \multicolumn{4}{c|}{EN~$\to$} & \multicolumn{4}{c}{\nonen{}~$\to$} \\
        & & & EL & UK & ID & SV & EL & UK & ID & SV \\
        \hline
        \multirow{3}{*}{\modelref{tgt_embed+lang-code}} & \multirow{3}{*}{Embed only} & None & -17 & -18 & -19 & -10 & -15 & -18 & -18 & -8 \\
        & & EN & -18 & -19 & -17 & -5 & -14 & -16 & -17 & -7 \\
        & & Proxy & -6 & -2 & -2 & -13 & -3 & +0 & -2 & -6 \\
        \hline
        \multirow{3}{*}{\modelref{tgt_embed+dec-only-last+lang-code}} & \multirow{3}{*}{\modelrefpar{tgt_embed+lang-code} + dec-last-layer} & None & -2 & -4 & -3 & +3 & -4 & -6 & -3 & -1 \\
        & & EN & +0 & -2 & -1 & +5 & -3 & -5 & -3 & +0 \\
        & & Proxy & +3 & +1 & -1 & +4 & +0 & +1 & -1 & +2 \\
        \hline
        \multirow{1}{*}{\modelref{tgt_embed+dec-only-last+enc-adapters-6-1024+lang-code}} & \modelrefpar{tgt_embed+dec-only-last+lang-code} + enc-adapters-last (d=1024) & None & -1 & -4 & +2 & -6 & -2 & -5 & -3 & -2 \\
        \hline
        \modelref{tgt_embed+adapters-430+lang-code} & Enc-adapters-all (dim=430) & None & +2 & -1 & -1 & +1 & -8 & +1 & +1 & -3 \\
        \hline
        \multirow{2}{*}{\modelref{tgt_embed+enc-adapters-6-1024+dec-adapters-690+lang-code}} & \modelrefpar{tgt_embed+lang-code} + enc-adapters-last (d=1024) & \multirow{2}{*}{None} & \multirow{2}{*}{-1} & \multirow{2}{*}{-1} & \multirow{2}{*}{1} & \multirow{2}{*}{-1} & \multirow{2}{*}{-2} & \multirow{2}{*}{-4} & \multirow{2}{*}{-1} & \multirow{2}{*}{-4} \\
        & + dec-adapters-all (d=690) & & & & & & \\
    \end{tabular}
    \caption{TED valid chrF delta ($\times1000$) of target-side incremental learning techniques with fixed language codes, compared to models with learned language codes. ``None'' corresponds to training and decoding without any language code. ``EN'' trains and decodes with the pre-trained (and frozen) ``to English'' language code. ``Proxy'' uses the closest pre-trained language code (RU for UK, BG for EL, DE for SV and VI for ID). This is an oracle, obtained by computing the Euclidean distance between trained language codes in \modelrefpar{retraining_4}.}
    \label{tab:lang_code_ablation_study}
\end{table}

\begin{table*}
    \centering
    \begin{tabular}{c@{\hspace{.2cm}}|c|ccc}
    ID & Model & $\to$EN & $\leftarrow$EN & \nonen{} \\
    \hline
    \modelref{paracrawl_en_centric_big_6_6} & Big 6-6 English-centric & .582 & .571 & .400 \\
    \modelref{paracrawl_en_centric} & Big 12-2 English-centric & \textbf{.587} & \textbf{.577} & .435 \\
    \modelref{paracrawl_multiparallel} & \modelrefpar{paracrawl_en_centric} + multi-parallel training & .583 & .573 & .486 \\
    \modelref{paracrawl_en_centric_pivot} & \modelrefpar{paracrawl_en_centric} + pivot through English & -- & -- & \textbf{.488} \\
    \hline
    \modelref{paracrawl_en_centric_retraining} & \modelrefpar{paracrawl_en_centric} + \threelang{} & .585 & .574 & .433 \\
    \modelref{paracrawl_retraining} & \modelrefpar{paracrawl_multiparallel} + \threelang{} & .580 & .569 & .486 \\
    \end{tabular}
    \caption{TED2020-valid chrF scores of the ParaCrawl/UNPC baselines.}
    \label{tab:paracrawl_baselines_valid_chrf}
\end{table*}

\begin{table*}
    \centering
    \begin{tabular}{c@{\hspace{.2cm}}|cH|ccc|ccc}
        \multirow{2}{*}{ID} & \multirow{2}{*}{Model} & \multirow{2}{*}{Params} & AR & RU & ZH & AR & RU & ZH \\
        & & & \multicolumn{3}{c|}{$\to$ EN} & \multicolumn{3}{c}{$\to$ \nonen{}} \\
        \hline
        \modelref{paracrawl_bilingual} & Bilingual baselines (pivot through English) & 193.2M & .499 & .460 & .430 & \textbf{.429} & .423 & .381 \\
        \modelref{paracrawl_en_centric_retraining} & English-centric \modelrefpar{paracrawl_en_centric} + \threelang{} & -- & .488 & \textbf{.480} & .430 & .372 & .385 & .337 \\
        \modelref{paracrawl_retraining} & Multi-parallel \modelrefpar{paracrawl_multiparallel} + \threelang{} & -- & .494 & .479 & .430 & .395 & .418 & .345 \\
        \modelref{paracrawl_retraining_pivot} & \modelrefpar{paracrawl_retraining} + pivot through English & -- & -- & -- & -- & .424 & \textbf{.433} & \textbf{.382} \\
        \hline
        \modelref{para_src_embed} & Only embed & 8.6M & .447 & .469 & .416 & .378 & .425 & .365 \\
        \setmodel{para_src_embed_no_lang_id} & \modelrefpar{para_src_embed} without lang ID filtering & 8.6M & .447 & .469 & .416 & .378 & .425 & .365 \\
        \modelref{para_src_embed+adapters_512} & \modelrefpar{para_src_embed} + enc-adapters-all (d=512) & 21.3M & \textbf{.502} & .478 & \textbf{.434} & .154 & .157 & .152 \\
        \modelref{para_src_embed+layer-1} & \modelrefpar{para_src_embed} + enc-first-layer & 21.2M & .491 & .474 & .428 & .154 & .168 & .152 \\
        \setmodel{para_src_embed+layer-1_no_lang_id} & \modelrefpar{para_src_embed+layer-1} without lang ID filtering & 21.2M & .488 & .474 & .427 & .154 & .158 & .151 \\
        \modelref{para_src_embed+layer-1_multipara} & \modelrefpar{para_src_embed+layer-1} + 20k lines per lang (BT) & 21.2M & .492 & .474 & .427 & .417 & \textbf{.433} & .376 \\
        \setmodel{para_src_embed+layer-1_multipara_no_lang_id} & \modelrefpar{para_src_embed+layer-1_multipara} without lang ID filtering & 21.2M & .433 & .457 & .385 & .148 & .158 & .148 \\
    \end{tabular}
    \\[.5cm]
    \begin{tabular}{c@{\hspace{.2cm}}|cH|ccc|ccc}
        \multirow{2}{*}{ID} & \multirow{2}{*}{Model} & \multirow{2}{*}{Params} & \multicolumn{3}{c|}{EN $\to$} & \multicolumn{3}{c}{\nonen{} $\to$} \\
        & & & AR & RU & ZH & AR & RU & ZH \\
        \hline
        \modelref{paracrawl_bilingual} & Bilingual baselines (pivot through English) & 193.2M & .423 & .437 & .187 & .358 & .400 & .156 \\
        \modelref{paracrawl_en_centric_retraining} & English-centric \modelrefpar{paracrawl_en_centric} + \threelang{} & -- & .412 & .439 & .179 & .295 & .384 & .115 \\
        \modelref{paracrawl_retraining} & Multi-parallel \modelrefpar{paracrawl_multiparallel} + \threelang{} & -- & .423 & .443 & .182 & .300 & .402 & .126 \\
        \modelref{paracrawl_retraining_pivot} & \modelrefpar{paracrawl_retraining} + pivot through English & -- & -- & -- & -- & .356 & .402 & .153 \\
        \hline
        \modelref{para_tgt_embed} & Only embed & 8.6M & .314 & .398 & .158 & .277 & .364 & .134 \\
        \setmodel{para_tgt_embed_no_lid} & \modelrefpar{para_tgt_embed} without lang ID filtering & 8.6M & .282 & .395 & .130 & .224 & .301 & .084 \\
        \modelref{para_tgt_adapters} & \modelrefpar{para_tgt_embed} + enc-adapters-last + dec-adapters-all (d=1024) & 14.9M & .417 & .437 & .187 & .348 & .397 & .153 \\
        \modelref{para_tgt_last_layer} & \modelrefpar{para_tgt_embed} + dec-last-layer & 25.4M & .412 & .441 & .187 & .348 & .401 & .154 \\
        \modelref{para_tgt_last_layer+enc_adapter} & \modelrefpar{para_tgt_last_layer} + enc-adapters-last (d=1024) & 27.5M & \textbf{.426} & \textbf{.446} & .192 & .356 & \textbf{.404} & .156 \\
        \modelref{para_tgt_last_layer+enc_adapter_no_lang_id} & \modelrefpar{para_tgt_last_layer+enc_adapter} without lang ID filtering & 27.5M & .312 & .363 & .107 & .072 & .185 & .037 \\
        \modelref{para_tgt_full_dec} & \modelrefpar{para_tgt_embed} + dec-all-layers & 42.2M & \textbf{.426} & \textbf{.446} & \textbf{.193} & \textbf{.357} & \textbf{.404} & \textbf{.159} \\
    \end{tabular}
    \caption{TED2020-valid chrF scores of the ParaCrawl/UNPC incrementally-trained models.}
    \label{tab:paracrawl_inc_valid_chrf}
\end{table*}

\begin{table}
    \centering
    \begin{tabular}{Hcc|cc|c}
        ID & & Source model & & Target model & chrF \\
        \hline
        & \modelref{paracrawl_bilingual} & \multicolumn{3}{c|}{Bilingual baselines (pivot through English)} & \textbf{.274} \\
        & \modelref{paracrawl_retraining} & \multicolumn{3}{c|}{Multi-parallel \modelrefpar{paracrawl_multiparallel} + \threelang{}} & .222 \\
        & \modelref{paracrawl_retraining_pivot} & \multicolumn{3}{c|}{\modelrefpar{paracrawl_retraining} + pivot through English} & .271 \\
        \hline
        & \multirow{2}{*}{\modelref{para_src_embed}} & \multirow{2}{*}{Only embed} & \modelref{para_tgt_last_layer} & Dec-last-layer & .250 \\
        & & & \modelref{para_tgt_last_layer+enc_adapter} & Dec-last-layer + enc-adapters-last (d=1024) & .252 \\
        \hline
        & \multirow{3}{*}{\modelref{para_src_embed+adapters_512}} &  \multirow{3}{*}{Enc-adapters-all (d=512)} & \modelref{para_tgt_last_layer} & Dec-last-layer & .246 \\
        & & & \modelref{para_tgt_last_layer+enc_adapter} & Dec-last-layer + enc-adapters-last (d=1024) & .251 \\
        & & & \modelref{para_tgt_last_layer+enc_adapter} & Pivot through English$^\star$ & \textbf{.274} \\
        \hline
        & \multirow{2}{*}{\modelref{para_src_embed+layer-1}} & \multirow{2}{*}{Enc-first-layer} & \modelref{para_tgt_last_layer} & Dec-last-layer & .229 \\
        & & & \modelref{para_tgt_last_layer+enc_adapter} & Dec-last-layer + enc-adapters-last (d=1024) & .234 \\
        \hline
        & \multirow{2}{*}{\modelref{para_src_embed+layer-1_multipara}} & \multirow{2}{*}{Enc-first-layer + 20k (BT)} & \modelref{para_tgt_last_layer} & Dec-last-layer & .247 \\
        & & & \modelref{para_tgt_last_layer+enc_adapter} & Dec-last-layer + enc-adapters-last (d=1024) & .251 \\
    \end{tabular}
    \vspace{-.15cm}
    \caption{TED2020-valid chrF scores of the ParaCrawl/UNPC models on \threelang{}$\to$\threelang{} (average over 6 directions) by combining source-language and target-language incrementally-trained parameters. ($\star$) instead of combining model parameters, translate to English with \modelrefpar{para_src_embed+layer-1}, then to the target language with \modelrefpar{para_tgt_last_layer+enc_adapter}.}
    \label{tab:paracrawl_source_target_inc_valid_chrf}
\end{table}

\begin{table}
    \centering
    \begin{tabular}{m{4.2cm}|p{10.5cm}}
        English-centric & Denotes a parallel corpus that only has alignments with English (i.e., 38 language pairs in our settings). Also denotes many-to-many models that are trained with such data. In this setting, translation between non-English language pairs is called ``zero-shot.'' \\
        \hline
        Multi-parallel & Denotes a parallel corpus that has alignments between all possible language pairs (380 in our case), and by extension, models that are trained with such data (AKA ``complete multilingual  NMT'', \citealp{freitag-firat-2020-complete}). \\
        \hline
        Enc-adapters-X ($d=N$) & Train adapter modules of bottleneck dimension $N$ after the encoder layer $X$. \\
        \hline
        Enc-adapters-all ($d=N$) & Train adapter modules of bottleneck dimension $N$ after all encoder layers. \\
        \hline
        Enc-layer-X & Fine-tune the $X$th Transformer encoder layer. \\
        \hline
        Enc-norm + enc-biases & Fine-tune the layer norm parameters and all the biases in the Transformer encoder. \\ 
        \hline
        + random embed init & Initialize the new language-specific embeddings at random, instead of initializing the embeddings of the known tokens with their previous values. \\
        \hline
        +~non-tied & Train separate target embeddings and output projection matrix (by default they are tied, i.e., they correspond to the same parameter). \\
        \hline
        + \multialigned{EL} & Train with multi-aligned EL data: not just paired with English, but with all of the 20 languages (2.41M lines pairs instead of 134k). \\
         \hline
        + \multialigned{EL} (BT) & Like above, but the non-EN data is obtained by translating the English side of the EL-EN corpus to the other 19 languages with \modelrefpar{en_centric}. For better comparison with the above method, we back-translate the same number of lines per language as in the multi-aligned EL corpus.\\
        \hline
        +~N lines per lang & Append to the EL-EN training data $N$ line pairs for each of the 19 non-English languages. \\
        \hline
        +~N lines per lang (BT) & Like above, but the $N$ line pairs per language are obtained by translating the English side of the EL-EN corpus with \modelrefpar{en_centric}. \\
        \hline
        
        \hline
        \nonen{} & 19 non-English languages in the initial model. As a column header, it means an average score over all 342 non-English translation directions. \\
        \hline
        $\to$~EN & Average score over all 19 \Xtoen{X} translation directions. \\
        \hline
        $\leftarrow$~EN & Average score over all 19 \entoX{Y} translation directions. \\
        \hline
        Re-training + $\{L_1, L_2\dots\}$ & Fine-tune the initial model with an updated BPE vocabulary and embedding matrix that include the new languages ($L_1$, $L_2$, etc.), and on the multi-aligned data of all the 20 initial languages plus the new ones.\\
        \hline
        (K) + $\{L_1, L_2\dots\}$ & Use the incremental training technique $K$, but on several languages at once ($L_1$, $L_2$, etc.) This means that a single shared BPE is trained for all these languages (whose size is multiplied by the number of languages) and the newly-trained parameters are shared between them. \\
    \end{tabular}
    \caption{Summary of the notations used in this paper.}
    \label{tab:notations}
\end{table}

\end{document}